\pdfoutput=1
\documentclass[10pt,logo,twocolumn,copyright]{nvidiatechreport}
\usepackage[numbers]{natbib}

\usepackage{nicefrac}
\usepackage{multirow}
\usepackage{multicol}
\usepackage{xspace}
\usepackage{adjustbox}
\usepackage{wrapfig}
\usepackage{dblfloatfix}
\usepackage{float}
\usepackage{makecell}
\usepackage{hhline}
\usepackage{array}
\usepackage{diagbox}
\usepackage{subcaption}

\usepackage[flushmargin]{footmisc}
\usepackage{mathtools}
\usepackage{xfp}
\usepackage{siunitx}
\usepackage{tikz}
\usepackage{pgfplots}
\pgfplotsset{compat=1.18}
\usepgfplotslibrary{groupplots,fillbetween}

\usepackage[capitalize,noabbrev]{cleveref}

\usepackage{etoolbox}
\AtBeginEnvironment{table}{\setlength{\parskip}{0pt}}
\AtBeginEnvironment{table*}{\setlength{\parskip}{0pt}}

\usepackage{amsmath,amsfonts,bm}

\def\eqref#1{equation~\ref{#1}}

\def\1{\bm{1}}

\def\vb{{\bm{b}}}
\def\vc{{\bm{c}}}

\def\vk{{\bm{k}}}

\def\vq{{\bm{q}}}

\def\vu{{\bm{u}}}
\def\vv{{\bm{v}}}

\def\vx{{\bm{x}}}
\def\vy{{\bm{y}}}

\def\mA{{\bm{A}}}

\def\mS{{\bm{S}}}

\DeclareMathAlphabet{\mathsfit}{\encodingdefault}{\sfdefault}{m}{sl}
\SetMathAlphabet{\mathsfit}{bold}{\encodingdefault}{\sfdefault}{bx}{n}

\newcommand{\R}{\mathbb{R}}

\newcommand{\softmax}{\mathrm{softmax}}

\theoremstyle{plain}

\theoremstyle{definition}

\theoremstyle{remark}

\title{Stateful Token Reduction for Long-Video Hybrid VLMs}

\author{Jindong Jiang* \quad
Amala Sanjay Deshmukh \quad
Kateryna Chumachenko \quad
Karan Sapra \quad
Zhiding Yu \quad
Guilin Liu \quad
Andrew Tao \quad
Pavlo Molchanov \quad
Jan Kautz \quad
Wonmin Byeon* \\~\\
\large{NVIDIA}}

\begin{abstract}

Token reduction accelerates long-video vision--language models (VLMs), but existing methods target Transformers, where reduction is treated as token pruning. We study token reduction in hybrid Mamba--Transformer VLMs and find that it is \emph{stateful}: Mamba layers maintain a recurrent state that accumulates information from earlier tokens, allowing discarded tokens to persist, so reduction behaves more like compression than dropping.
We support this view with a representation-based probing method measuring how much information from discarded tokens is retained, and analyze layer-wise sparsity and cross-layer importance stability. Our findings show importance is sparse within layers but unstable across layers, making aggressive early pruning unreliable while hybrids remain robust to later reduction.
Motivated by this, we propose a hybrid-aware token reduction framework with a low-to-high progressive schedule and a unified query-conditioned importance score for attention and Mamba layers. For Mamba, excluding the position-dependent decay from the recurrence produces a stronger selection signal. 
Across long-video benchmarks, our method achieves $3.8$--$4.2\times$ prefilling speedups at a 25\% token budget while maintaining near-baseline accuracy and improving with light finetuning. Hybrid models benefit from aggressive reduction, improving both efficiency and accuracy, whereas Transformers exhibit the standard trade-off. Our method also outperforms prior baselines on the same hybrid backbone and combines effectively with visual redundancy reduction methods.

\end{abstract}

\begin{document}

\maketitle

\section{Introduction}
\label{sec:intro}

Video-based vision--language models (VLMs) are moving toward long-horizon video understanding tasks. However, long-video VLMs process thousands of visual tokens, making inference expensive, especially during prefilling. Prior work shows that many tokens are redundant and can be pruned or merged, but aggressive reduction often degrades accuracy, particularly when applied early, where errors cannot be corrected in later layers \cite{chen2024imageworth12tokens,zhang2025sparsevlmvisualtokensparsification,xing2024pyramiddrop}. This becomes more challenging for long videos, where token relevance varies across depth and time \cite{zhang2025beyond,wang2025metok,zhang2025flexselect}.

Existing token-reduction methods are primarily designed for Transformer architectures, where reduction is naturally interpreted as token pruning. However, recent VLMs increasingly adopt hybrid Mamba--Transformer architectures \cite{nvidia2025nvidianemotronnano2,nvidia2025nvidianemotronnanov2}, in which a substantial fraction of layers are state-space (Mamba) blocks. These architectures propagate information differently, but their implications for token reduction remain underexplored.

In this work, we study token reduction in hybrid VLMs and find it exhibits \emph{stateful} behavior. In contrast to Transformers, where removed tokens no longer influence subsequent layers, 
Mamba layers maintain a recurrent state that accumulates information from earlier tokens and carries it across layers.
As a result, information from discarded tokens can persist in the state, and reduction behaves more like compression than dropping.

We show a representation-based probe measuring how much 
information from discarded tokens is preserved in retained tokens, providing empirical evidence for the stateful behavior of hybrid models. This observation has implications for reduction design. Complementing this, we analyze layer-wise sparsity and cross-layer importance stability, and find that while importance is sparse within each layer, it is unstable across layers, making aggressive early pruning unreliable. At the same time, hybrid models show stronger tolerance to reduction, consistent with information being absorbed into the recurrent state. These findings motivate a progressive reduction strategy that preserves more tokens in early layers and prunes more aggressively later. 

Based on this analysis, we propose a hybrid-aware token reduction framework. We adopt a low-to-high progressive schedule and introduce a unified query-conditioned importance score for both attention and Mamba layers. For attention layers, importance is derived from text-to-vision attention. For Mamba layers, we derive an attention-like score from the selective state-space recurrence and show that the content-alignment term alone is a more effective signal than the full recurrent weight, which includes position-dependent decay.

Experiments on long-video benchmarks (VideoMME~\cite{fu2025video}, LongVideoBench~\cite{wu2024longvideobench}, and LVBench~\cite{wang2025lvbench}) show that the proposed method achieves substantial prefilling and end-to-end speedups. Notably, hybrid models maintain or improve accuracy under aggressive token reduction, whereas Transformer models exhibit the standard speed--accuracy trade-off. Ours also outperforms existing token-reduction baselines on the same hybrid backbone and combines effectively with visual redundancy reduction methods.

\paragraph{Contributions.}
(i) We study token reduction in Mamba--Transformer hybrid VLMs and identify its stateful behavior, supported by a representation-based probing analysis. 
(ii) We propose a hybrid-aware token reduction framework with a progressive schedule and a unified query-conditioned scoring method across attention and Mamba layers, and show that excluding cumulative decay is critical for Mamba-based selection. 
(iii) We demonstrate strong speed--quality trade-offs on long-video benchmarks, with hybrid models benefiting from token reduction more than Transformer baselines.

The rest of the paper is organized as follows. \Cref{sec:analysis} presents the empirical observations that motivate our design. \Cref{sec:method} describes the scoring rule and the progressive schedule. \Cref{sec:experiment} reports the main results, ablations, and efficiency analysis. A broader discussion of related work is provided in \cref{sec:related} and \cref{sec:related_appendix} (Appendix).

\section{Observations}
\label{sec:analysis}

We compare a Transformer VLM (Qwen3-VL) and a hybrid Mamba--Transformer VLM (Nemotron-Nano-V2 VL) along three properties that motivate the method in \cref{sec:method}: \emph{(i)~stateful compression}, \emph{(ii)~layer-wise sparsity}, and \emph{(iii)~cross-layer importance stability}. We use the query-conditioned token-importance score whose precise definition is deferred to \cref{sec:token_importance}. 

\subsection{Stateful Compression}
\label{sec:obs_stateful}
We hypothesize that consistency and reduction tolerance are driven by different mechanisms. In standard Transformers, attention is memoryless, so when the tokens are removed, their information is effectively \emph{dropped}. Once pruned, these tokens cannot influence later layers~\cite{xiao2025sliding}. Hybrid models behave differently. They interleave attention with Mamba layers that maintain a recurrent state $\mS_t$. This allows token reduction to act more like \emph{compression} rather than dropping. Even if a token is discarded, its information can persist in the state and be propagated to the remaining tokens.

\paragraph{Probing Stateful Compression.}
We test this idea using a representation probe. First, we run a full forward pass and select the top-$K$ tokens based on their importance at layer $\ell$. Then, we run a second forward pass where only these $K$ tokens are kept from layer~$1$ onward. At layer $\ell$, we compare the hidden states of the retained tokens between the two passes using angular distance. If the model simply drops tokens, the two representations should be very similar. In contrast, a larger angular distance indicates that earlier layers have already absorbed information from the discarded tokens into the retained ones. \cref{fig:stateful_compression_probe} shows results across two retention budgets equivalent to $32$ and $16$ frames, three depths ($\sim$11\%, $\sim$26\%, $\sim$41\%), and up to $512$ frames. The hybrid model consistently shows larger angular distances than the Transformer (e.g., $45.2^\circ$ vs.\ $22.2^\circ$ at 41\% depth, $32$-frame budget, $256$ frames). Moreover, this gap increases with both depth and number of frames, supporting the idea that hybrid models perform stateful compression rather than simple token dropping.

\subsection{Layer Sparsity Statistics}
\label{sec:obs_sparsity}

\begin{figure*}[t]
\centering
\begin{subfigure}[b]{0.43\textwidth}
\vspace{0pt}
  \centering
      \includegraphics[width=1.0\linewidth]{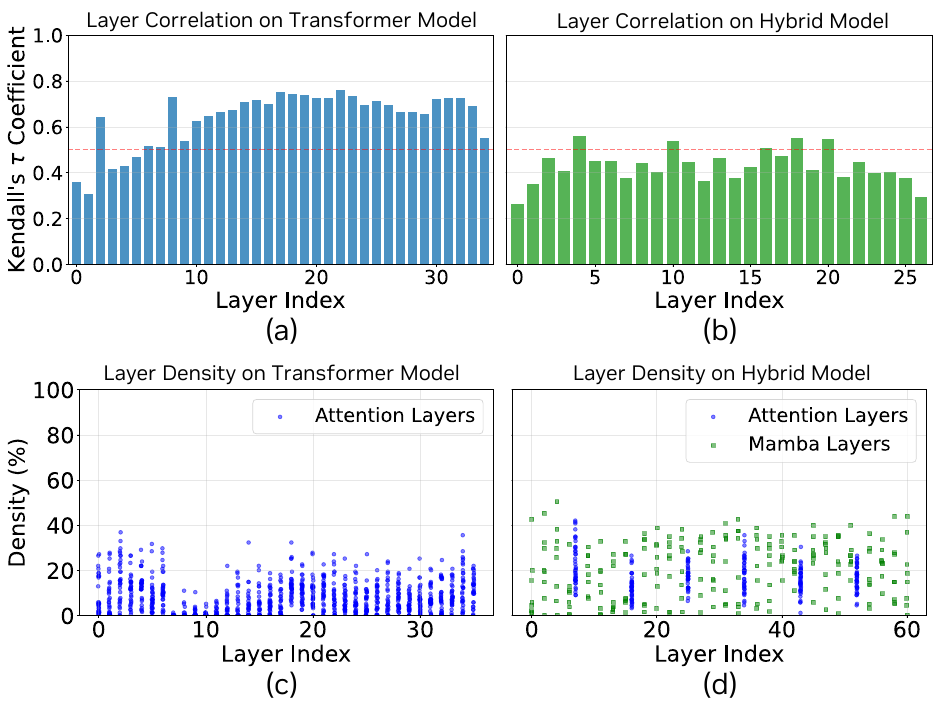}
      \caption{Layer-wise attention density and cross-layer rank stability of token importance.}
  \label{fig:sparsity_analysis}
\end{subfigure}
\hfill
\begin{subfigure}[b]{0.56\textwidth}
\vspace{0pt}
\centering
\definecolor{cbHybrid}{HTML}{D55E00}    
\definecolor{cbTrans}{HTML}{0072B2}     
\definecolor{gridGray}{HTML}{D9D9D9}

\resizebox{\linewidth}{!}{%
\begin{tikzpicture}
\pgfplotsset{
  compat=1.18,
  every axis/.style={
    width=2.13in,
    height=1.65in,
    xmode=log,
    log basis x=2,
    log ticks with fixed point,
    xticklabel style={/pgf/number format/.cd, fixed, precision=0, /tikz/.cd, inner sep=0.5pt},
    ymin=0, ymax=65,
    ytick={0,15,30,45,60},
    enlargelimits=false,
    axis line style={black!60, line width=0.4pt},
    tick style={black!60, line width=0.4pt},
    tick label style={font=\scriptsize},
    label style={font=\footnotesize},
    title style={font=\footnotesize\bfseries, yshift=-6pt},
    grid=major,
    major grid style={draw=gridGray, line width=0.25pt},
  },
  hybrid/.style={
    color=cbHybrid, line width=0.9pt,
    mark=*, mark size=1.4pt,
    mark options={solid, fill=cbHybrid, draw=white, line width=0.25pt}
  },
  trans/.style={
    color=cbTrans, dashed, dash pattern=on 3pt off 2pt, line width=0.85pt,
    mark=square*, mark size=1.3pt,
    mark options={solid, fill=cbTrans, draw=white, line width=0.25pt}
  },
  gapfill/.style={
    fill=cbHybrid, fill opacity=0.14, draw=none, forget plot
  },
}

\tikzset{
  inpanellegend/.style={
    anchor=north west,
    fill=white,
    fill opacity=0.88,
    text opacity=1,
    inner sep=1.5pt,
    font=\scriptsize,
    align=left
  },
  endpointgap/.style={
    black!60,
    line width=0.45pt
  },
  endpointguide/.style={
    black!35,
    densely dotted,
    line width=0.35pt
  },
  endpointgaplabel/.style={
    anchor=south east,
    xshift=-2pt,
    yshift=1pt,
    inner sep=0.8pt,
    font=\scriptsize\bfseries,
    text=cbHybrid!75!black,
    fill=white,
    fill opacity=0.82,
    text opacity=1
  },
}
\newcommand{\drawendpointgap}[4]{%
  \draw[endpointguide] (axis cs:#1,0) -- (axis cs:#1,#2);
  \draw[endpointgap] (axis cs:#1,#2) -- (axis cs:#1,#3);
  \draw[endpointgap] (axis cs:#1,#2) ++(-1.2pt,0) -- ++(2.4pt,0);
  \draw[endpointgap] (axis cs:#1,#3) ++(-1.2pt,0) -- ++(2.4pt,0);
  \node[endpointgaplabel] at (axis cs:#1,#3) {$+#4\%$};
}

\begin{groupplot}[
    group style={
        group size=3 by 2,
        horizontal sep=0.10in,
        vertical sep=0.30in,
    },
]

\nextgroupplot[
    title={$\sim$11\% model depth},
    xtick={64,128,256,512},
    xmin=55, xmax=600,
]
\addplot[name path=Hr1A, draw=none, forget plot] coordinates {(64,4.6) (128,7.3) (256,10.0) (512,14.4)};
\addplot[name path=Tr1A, draw=none, forget plot] coordinates {(64,2.9) (128,4.0) (256,4.5) (512,7.6)};
\addplot[gapfill] fill between[of=Hr1A and Tr1A];
\addplot[hybrid] coordinates {(64,4.6) (128,7.3) (256,10.0) (512,14.4)};
\addplot[trans]  coordinates {(64,2.9) (128,4.0) (256,4.5) (512,7.6)};
\drawendpointgap{512}{7.6}{14.4}{89}
\node[inpanellegend] at (rel axis cs:0.04,0.96) {%
    \tikz[baseline=-0.5ex]{%
        \draw[cbHybrid, line width=0.9pt] (0,0)--(0.18in,0);
    }\,Hybrid\\[-1pt]%
    \tikz[baseline=-0.5ex]{%
        \draw[cbTrans, dashed, dash pattern=on 3pt off 2pt, line width=0.85pt] (0,0)--(0.18in,0);
    }\,Transformer
};

\nextgroupplot[
    title={$\sim$26\% model depth},
    xtick={64,128,256,512},
    yticklabels=\empty,
    xmin=55, xmax=600,
]
\addplot[name path=Hr1B, draw=none, forget plot] coordinates {(64,13.0) (128,22.1) (256,28.8) (512,35.6)};
\addplot[name path=Tr1B, draw=none, forget plot] coordinates {(64,5.1) (128,7.3) (256,9.4) (512,20.0)};
\addplot[gapfill] fill between[of=Hr1B and Tr1B];
\addplot[hybrid] coordinates {(64,13.0) (128,22.1) (256,28.8) (512,35.6)};
\addplot[trans]  coordinates {(64,5.1) (128,7.3) (256,9.4) (512,20.0)};
\drawendpointgap{512}{20.0}{35.6}{78}

\nextgroupplot[
    title={$\sim$41\% model depth},
    xtick={64,128,256,512},
    yticklabels=\empty,
    xmin=55, xmax=600,
]
\addplot[name path=Hr1C, draw=none, forget plot] coordinates {(64,20.8) (128,35.0) (256,45.2) (512,52.8)};
\addplot[name path=Tr1C, draw=none, forget plot] coordinates {(64,11.8) (128,17.7) (256,22.2) (512,33.8)};
\addplot[gapfill] fill between[of=Hr1C and Tr1C];
\addplot[hybrid] coordinates {(64,20.8) (128,35.0) (256,45.2) (512,52.8)};
\addplot[trans]  coordinates {(64,11.8) (128,17.7) (256,22.2) (512,33.8)};
\drawendpointgap{512}{33.8}{52.8}{56}

\nextgroupplot[
    xlabel={\# Frames},
    xtick={32,64,128,256},
    xmin=28, xmax=300,
]
\addplot[name path=Hr2A, draw=none, forget plot] coordinates {(32,2.8) (64,5.8) (128,8.4) (256,10.8)};
\addplot[name path=Tr2A, draw=none, forget plot] coordinates {(32,1.8) (64,3.3) (128,4.3) (256,4.8)};
\addplot[gapfill] fill between[of=Hr2A and Tr2A];
\addplot[hybrid] coordinates {(32,2.8) (64,5.8) (128,8.4) (256,10.8)};
\addplot[trans]  coordinates {(32,1.8) (64,3.3) (128,4.3) (256,4.8)};
\drawendpointgap{256}{4.8}{10.8}{125}

\nextgroupplot[
    xlabel={\# Frames},
    xtick={32,64,128,256},
    yticklabels=\empty,
    xmin=28, xmax=300,
]
\addplot[name path=Hr2B, draw=none, forget plot] coordinates {(32,8.3) (64,16.2) (128,24.9) (256,31.2)};
\addplot[name path=Tr2B, draw=none, forget plot] coordinates {(32,3.4) (64,6.1) (128,8.5) (256,10.5)};
\addplot[gapfill] fill between[of=Hr2B and Tr2B];
\addplot[hybrid] coordinates {(32,8.3) (64,16.2) (128,24.9) (256,31.2)};
\addplot[trans]  coordinates {(32,3.4) (64,6.1) (128,8.5) (256,10.5)};
\drawendpointgap{256}{10.5}{31.2}{197}

\nextgroupplot[
    xlabel={\# Frames},
    xtick={32,64,128,256},
    yticklabels=\empty,
    xmin=28, xmax=300,
]
\addplot[name path=Hr2C, draw=none, forget plot] coordinates {(32,13.6) (64,24.2) (128,38.0) (256,47.3)};
\addplot[name path=Tr2C, draw=none, forget plot] coordinates {(32,8.3) (64,13.2) (128,19.9) (256,24.7)};
\addplot[gapfill] fill between[of=Hr2C and Tr2C];
\addplot[hybrid] coordinates {(32,13.6) (64,24.2) (128,38.0) (256,47.3)};
\addplot[trans]  coordinates {(32,8.3) (64,13.2) (128,19.9) (256,24.7)};
\drawendpointgap{256}{24.7}{47.3}{91}

\end{groupplot}

\node[rotate=90, anchor=center, font=\footnotesize\bfseries]
     at ($(group c1r1.west)+(-0.22in,0)$) {32-Frame Budget};
\node[rotate=90, anchor=center, font=\footnotesize\bfseries]
     at ($(group c1r2.west)+(-0.22in,0)$) {16-Frame Budget};

\end{tikzpicture}
}
\caption{Stateful compression probing via angular distance of the hidden representations between full and the top-$K$ retained tokens.}
\label{fig:stateful_compression_probe}

\end{subfigure}
\caption{\textbf{Token importance structure and compression behavior in Hybrid vs Transformer models.} \textbf{(i)} \textbf{Top:} Kendall's $\tau$~\citep{kendall1938new} between the importance rankings of consecutive layers, for the Transformer~(a) and the hybrid~(b); values around 0.5 and below indicate low rank consistency. \textbf{Bottom:} layer-wise density (\%) of importance scores for the Transformer~(c) and the hybrid~(d), where each dot is one attention head (blue) or one Mamba group (green); lower density indicates higher sparsity. \textbf{(ii)} Angular distance between hidden states of retained tokens, quantifies how much information from discarded tokens is absorbed. Larger distances indicate more information absorbed from the discarded tokens, and small distance means the representations are almost unchanged, implying little additional information was incorporated. The Hybrid model consistently shows greater compression, with the gap widening at deeper layers and higher frame counts. }
\vspace{-0.5cm}
\end{figure*}

\cref{fig:sparsity_analysis} (bottom) plots the fraction of tokens that account for $80\%$ of the importance mass at each layer; each dot is one attention head (blue) or one Mamba group (green). The majority fall below $40\%$ in both architectures, with average density of $10.9\%$ in the Transformer and $30.6\%$ in the hybrid, suggesting that token reduction is feasible in principle for both. Transformer attention is sparser than Mamba on average, and within the hybrid, attention is also sparser than Mamba groups.

Sparsity tells us \emph{which} tokens matter at a given layer, but not whether the \emph{same} tokens remain important across layers. \cref{fig:sparsity_analysis} (top) reports Kendall's $\tau$ between consecutive layers' importance rankings; values below $0.5$ mean more than $25\%$ of token pairs reverse their relative ordering. Early-layer correlations are unstable in both architectures: the Transformer stabilizes only after roughly one-third of network depth, while the hybrid remains consistently low and on average lower than the Transformer's. The implication is that single-shot pruning at the very first layer is risky and a depth-aware schedule is preferable in either case.



\paragraph{Takeaways.}
The stronger stateful-compression effect in hybrids both \emph{permits} and \emph{rewards} preserving more tokens in early layers, so that the recurrent state can absorb information before later layers prune more aggressively. Together with~(ii) and~(iii), this motivates the low-to-high progressive schedule in \cref{sec:method}, which we expect to benefit hybrids in particular.

\section{Method}
\label{sec:method}

\subsection{Query-Conditioned Token Importance}
\label{sec:token_importance}

Consider a multimodal sequence with $M$ text (query) tokens and $N$ visual tokens. For long videos, $N$ can exceed 10K, creating computational bottlenecks, while $M$ is typically much smaller (e.g., $\sim$100). 
Let $\vu_1^{(\ell)}, \ldots, \vu_M^{(\ell)}$ and $\vv_1^{(\ell)}, \ldots, \vv_N^{(\ell)}$ denote the hidden states of text and visual tokens at layer $\ell$. 
Our goal is to define a \emph{query-conditioned importance score} $s_i^{(\ell)} \in \R$ for each visual token to measure relevance to the query at layer $\ell$ and enable ranking and pruning: given budget $K < N$, we retain the top-$K$ tokens by importance.

\paragraph{Attention Layers.}
For attention layers, importance follows naturally from the attention mechanism. Let $\vq_m^{(h)}$ and $\vk_i^{(h)}$ denote the query and key projections of $\vu_m^{(\ell)}$ and $\vv_i^{(\ell)}$ at head $h$ (we omit the layer index for brevity). We compute the softmax-normalized attention weights from text to visual tokens and aggregate:
$s_i^{(\ell, \text{attn})} = \frac{1}{MH} \sum_{m,h} \softmax_i\left( \frac{\vq_m^{(h)} \cdot \vk_i^{(h)}}{\sqrt{d_h}} \right),$ 
where $H$ is the number of heads, and $d_h$ is the head dimension.

\paragraph{State-Space (Mamba) Layers.}
For Mamba layers, we derive an implicit attention proxy from the selective state-space recurrence~\citep{dao2024transformersssmsgeneralizedmodels}. 
Each head in Mamba operates independently on input $\vx_t \in \R^{p}$ (where $p$ is the head dimension) with hidden state $\mS_t \in \R^{p \times n}$ (where $n$ is the state dimension). The state evolves as
$\mS_t = \bar{\mA}_t \mS_{t-1} + \vx_t \bar{\vb}_t^\top$, 
with output $\vy_t = \mS_t \vc_t$, 
where $\bar{\mA}_t$ is the discretized state transition matrix with entries in $(0,1)$ acting as a decay factor; $\bar{\vb}_t = \Delta_t \vb_t \in \R^n$ is the effective input projection combining the discretization step $\Delta_t > 0$ with the input-dependent projection $\vb_t$; and $\vc_t \in \R^n$ is the output projection. Mamba groups heads such that $\vb_t$ and $\vc_t$ are shared within each group.
Unrolling the recurrence reveals an attention-like structure $\vy_t = \sum_{j=1}^{t} w_{t,j} \, \vx_j$, where $w_{t,j}$ is a scalar weight quantifying the contribution of token $j$ to the output at position $t$:
\begin{equation}
w_{t,j} = \left( \prod_{u=j+1}^{t} \bar{\mA}_u \right) \bar{\vb}_j^\top \vc_t.
\label{eq:ssm_weight}
\end{equation}
This weight depends on two factors: (1) a \emph{content alignment} $\bar{\vb}_j^\top \vc_t$, where $\bar{\vb}_j$ and $\vc_t$ act analogously to keys and queries~\citep{katharopoulos2020transformers,deltaformer}, and (2) a \emph{cumulative decay} $\prod_u \bar{\mA}_u$ that monotonically attenuates older positions. We provide a detailed derivation of this attention connection in \cref{sec:appendix_linear_attention_connection}.

\paragraph{Why the Full Mamba Dynamic Fails.}
Given that \cref{eq:ssm_weight} is the exact contribution of visual token $j$ to the output at text position $t$, the most natural ``attention-like'' importance score is simply $|w_{t,j}|$, in direct analogy to softmax attention scores. We refer to this score as the \emph{Full Mamba Dynamic} because it preserves the full selective-scan dynamics, content alignment together with positional decay. However, we find that this choice performs substantially worse than what we propose below. Concretely, applying the Full Mamba Dynamic at $\sim$25\% token budget drops the average accuracy by up to $8.1$ points relative to the no-reduction baseline, depending on which Mamba layers participate in selection (see \cref{tab:scoring_ablation} in \cref{sec:experiment}). The cause is structural: 
the diagonal entries of $\bar{\mA}_u$ in Mamba layers are bounded in $(0,1)$ and their cumulative product $\prod_u \bar{\mA}_u$ can decay sharply with distance, so $w_{t,j}$ assigns near-zero weight to early visual tokens in some layers regardless of how relevant their content is to the query. We visualize this effect in some layers in \cref{fig:cross_attn_heatmap} (\cref{sec:appendix_token_selection}).

\paragraph{Decoupling Content from Position.}
This failure mode suggests separating the \emph{data-dependent} content alignment from the \emph{position-dependent} decay, and using only the former for token selection. We refer to the resulting score as the \emph{Decoupled Content Alignment}, defined for Mamba layers as
\begin{equation}
s_i^{(\ell, \text{ssm})} = \frac{1}{MG} \sum_{m,g} \left| \bar{\vb}_i^{(g)\top} \vc_m^{(g)} \right|,
\label{eq:ssm_importance}
\end{equation}
averaging over $M$ query (text) positions and $G$ Mamba groups, where $\bar{\vb}_i$ and $\vc_m$ act as key and query projections whose dot product measures alignment between visual token $i$ and text token $m$, with the gating factor $\Delta_i$ absorbed into $\bar{\vb}_i$. Despite its simplicity, this decomposition is, to our knowledge, the first explicit identification of which component of the Mamba recurrence is suitable for content-based token scoring and which component should be excluded; we validate this empirically against the Full Mamba Dynamic in \cref{tab:scoring_ablation}.

\paragraph{Token Selection.}
Given scores $\{s_i\}_{i=1}^{N}$, we select the top-$K$ tokens while preserving temporal order. The budget $K$ may vary across layers according to a reduction schedule, which we describe next, motivated by the observations in \cref{sec:analysis}.

\subsection{Reduction Scheduling}
\label{sec:reduction_strategy}

\begin{figure*}[t]
  \centering
  \begin{subfigure}[b]{0.45\textwidth}
  \centering
  \includegraphics[width=0.9\linewidth]{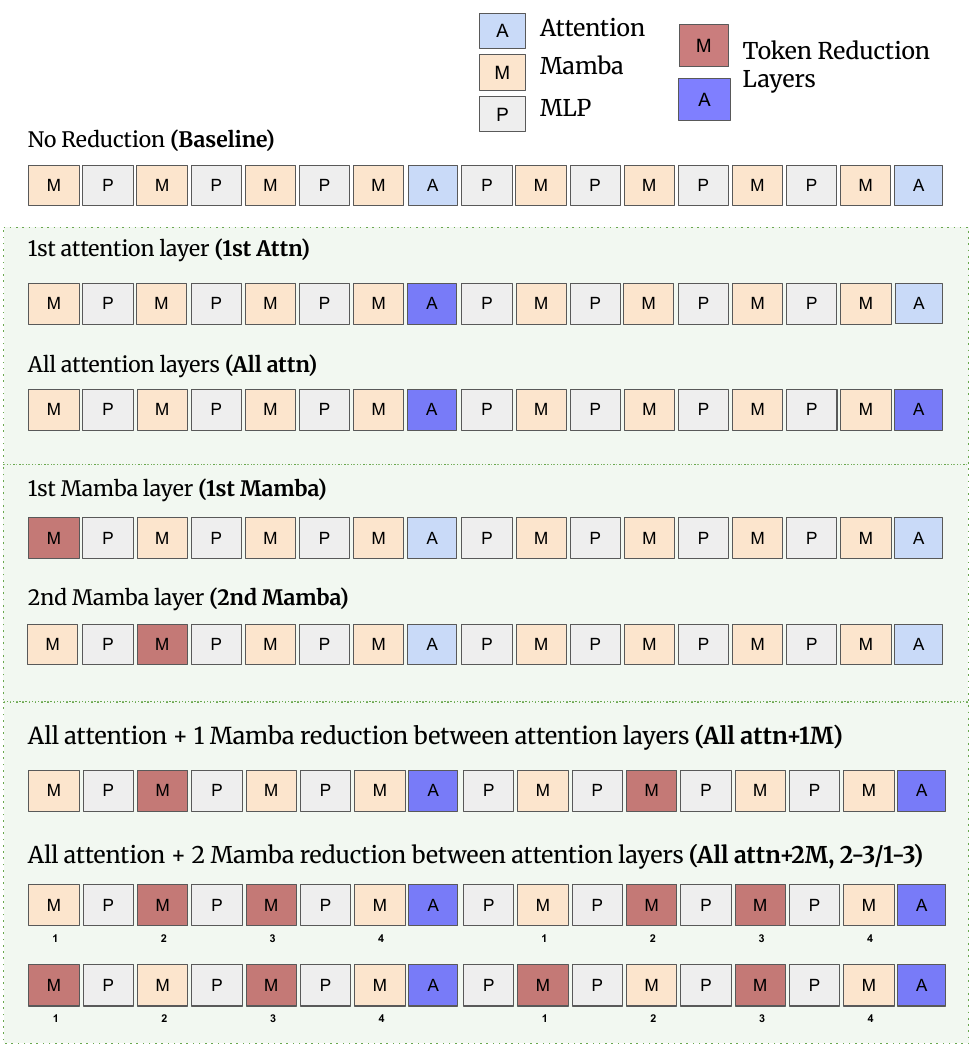}
  \caption{{Hybrid Token Reduction Patterns}}
  \label{fig:hybrid_patterns}
  \end{subfigure}
  \hfill
  \begin{subfigure}[b]{0.53\textwidth}
  \centering
  \includegraphics[width=0.9\linewidth]{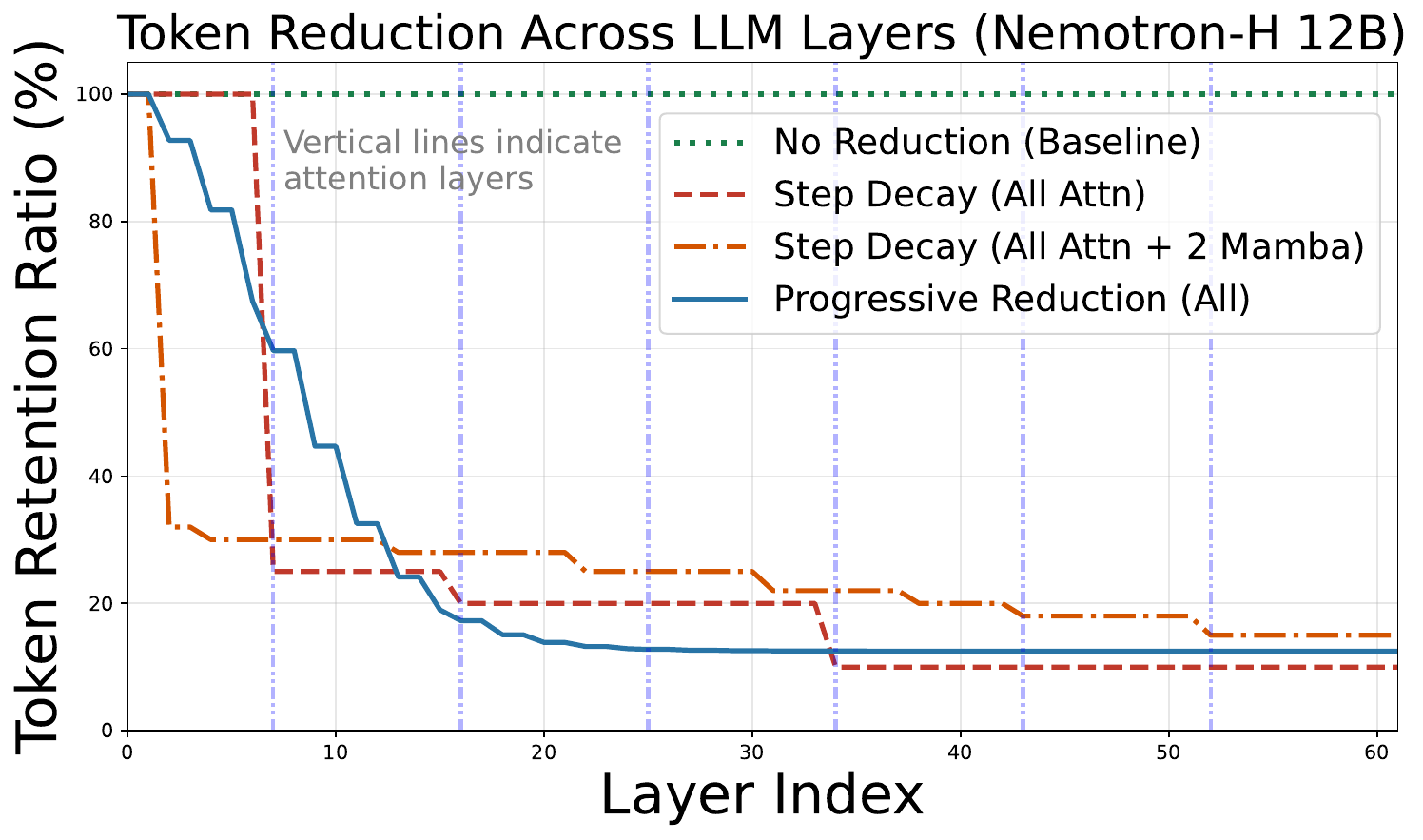}
  \caption{{Hybrid reduction schedules}}
  \label{fig:token_reduction_schedules}
  \end{subfigure}
\caption{\textbf{Token-reduction patterns and schedules used for the hybrid model.} \textbf{(i)} Layer types (Mamba/MLP/Attention) and reduction locations for baseline, attention-only, Mamba-only, and hybrid schedules (\texttt{All attn+1M}/\texttt{All attn+2M}). \textbf{(ii)} We visualize how the token retention ratio changes with depth: (1) no reduction, (2) step decay at attention layers, (3) step decay at attention + Mamba layers, and (4) progressive low-to-high reduction applied throughout the network. Vertical dotted lines mark attention layer locations.}
\vspace{-0.5cm}
\end{figure*}

We investigate how different reduction schedules interact with hybrid architectures, exploring a design space along two key dimensions: (1) \emph{where} to apply reduction, i.e., which layer types; and (2) \emph{when} to reduce, i.e., early versus progressive. \cref{fig:hybrid_patterns} illustrates the schedules we consider.

\paragraph{Layer Type Selection.}
Hybrid models interleave attention and Mamba layers (see \cref{fig:hybrid_patterns}), raising the question of where to apply reduction. We consider three options: \emph{attention-only}, where only attention layers compute importance and prune tokens; \emph{attention + Mamba}, which inserts one or two Mamba reductions between attention layers; and \emph{all layers}, where reduction is applied throughout the model. Attention-only is simpler and leverages attention sparsity, while including Mamba enables finer-grained pruning but requires computing the implicit importance scores in \cref{eq:ssm_importance}.

\paragraph{Early versus Progressive Reduction.}

A second design choice is whether to reduce tokens early or progressively. \emph{Early reduction} prunes tokens once at the LLM input using early-layer importance scores~\citep{khaki2025sparsevila}, while \emph{progressive reduction} retains more tokens in shallow layers and gradually reduces the budget with depth. Motivated by our analysis (\cref{sec:analysis}), i.e., early importance is unreliable, and Mamba layers can absorb information from discarded tokens, we adopt a low-to-high progressive schedule.
We parameterize the retention ratio $r^{(\ell)}$ as a monotonic decay from $r_{\text{start}}$ to $r_{\text{end}}$. We consider (1) \emph{step decay}, which reduces tokens in stages, and (2) \emph{sigmoid decay}~\citep{zhao2025accelerating}, which enables smooth, late-stage reduction via a midpoint-controlled curve, defined as $r^{(\ell)} = r_{\text{end}} + (r_{\text{start}} - r_{\text{end}}) \cdot \sigma(-k(\ell/L - x_0))$, where $L$ is the number of layers, $k$ controls steepness, and $x_0$ the midpoint. We fix $k=20$ and vary $x_0$ to control reduction position. Across target budgets (25\%, 35\%, 50\%), preserving more tokens in early layers matters more than the specific decay form. The reduction curve examples are shown in \cref{fig:token_reduction_schedules}.

We evaluate these scheduling and layer-type trade-offs empirically in \cref{sec:experiment}.

\section{Experiments}
\label{sec:experiment}

\subsection{Experimental Setup}
We use SigLIP2~\cite{tschannen2025siglip} (384$\times$384) to produce 576 patch tokens, pooled into 144 tokens/frame via an MLP projector. We compare a dense Transformer (Qwen3-VL 8B~\cite{bai2025qwen3vltechnicalreport}) and a Mamba–Transformer hybrid (Nemotron-Nano-V2 VL 12B~\cite{nvidia2025nvidianemotronnanov2}) under the same vision encoder and training pipeline. Training has four stages—alignment, SFT, long-context finetuning, and token-reduction finetuning—with frames increasing from 32$\to$64$\to$128 (96 for the no-reduction baseline). We align using 95K image–text pairs~\cite{jiang2025storm}, perform SFT on $\sim$12.5M mixed data, finetune on LLaVA-Video~\cite{zhang2024video}, and finally train with token reduction on EAGLE-Video-110K~\cite{chen2025eagle}. 
We evaluate on three long-context video benchmarks: VideoMME~\cite{fu2025video}, LongVideoBench (LongVB)~\cite{wu2024longvideobench}, and LVBench~\cite{wang2025lvbench}. VideoMME is reported without subtitles to isolate visual reasoning, while LongVB and LVBench focus on increasingly long videos (up to $\sim$1–2 hours) for long-horizon understanding.
Full architectural, training, and evaluation benchmark details are provided in the appendix~\ref{sec:appendix_experimental_setup}.


\paragraph{Reduction Schedules.}

To systematically evaluate the design choices discussed in \cref{sec:reduction_strategy}, we first test target token budgets of approximately 25\%, 35\%, and 50\% under an \emph{attention-only} reduction setting, which isolates the effect of budget on accuracy and latency (\cref{tab:compression_rate}). We then fix the budget to 25\% and compare different layer subsets, i.e., \emph{attention-only}, \emph{Mamba-only}, and \emph{attention+Mamba}, 
to study how reduction interacts with layer type (\cref{tab:token_reduction_hybrid}). For \emph{Mamba-only}, we additionally test early-layer-only reduction (1st vs.\ 2nd Mamba) to examine the sparsity/stability effects observed by the analysis in \cref{sec:analysis}. For \emph{attention-only}, we compare pruning at a single attention layer (1st attention) vs. all attention layers. Finally, to study reduction granularity, we evaluate intermediate hybrid configurations (\emph{attention + 1 or 2 Mamba}) that interpolate between attention-only and all-layer reduction. We visualize the layer-wise schedules and hybrid layer-patterns used in our experiments in \cref{fig:token_reduction_schedules} and \cref{fig:hybrid_patterns}. We report all schedule parameters in \cref{sec:appendix_reduction_schedules}. 


\newcolumntype{C}[1]{>{\centering\arraybackslash}m{#1}}

\definecolor{hlrow}{RGB}{248,216,216} 
\definecolor{posblue}{RGB}{0,60,200}
\definecolor{negred}{RGB}{200,0,0}

\providecommand{\pos}[1]{\textcolor{posblue}{#1}}
\providecommand{\nega}[1]{\textcolor{negred}{#1}}


\providecommand{\hybvaldelta}[2]{%
  \edef\dlt{\fpeval{round(#1-(#2),2)}}%
  \mbox{\num{#1}\,%
  \ifdim\dlt pt=0pt
    \pos{(+\num[round-mode=places,round-precision=2]{0})}%
  \else
    \ifdim\dlt pt>0pt
      \pos{(+\num[round-mode=places,round-precision=2]{\dlt})}%
    \else
      \nega{(-\num[round-mode=places,round-precision=2]{\fpeval{-(\dlt)}})}%
    \fi
  \fi}%
}

\providecommand{\ttftdelta}[2]{%
  \edef\val{\fpeval{round(#1,2)}}%
  \edef\dlt{\fpeval{round(#1-(#2),2)}}%
  \val\ %
  \ifnum\pdfstrcmp{\dlt}{0}=0
    \pos{(+0.00)}%
  \else
    \ifdim\dlt pt<0pt
      \pos{(\dlt)}
    \else
      \nega{(+\dlt)}
    \fi
  \fi
}

\providecommand{\hybttftfactor}[2]{%
  \edef\val{\fpeval{round(#1,2)}}%
  \edef\fac{\fpeval{round((#2)/(#1),1)}}%
  \mbox{\num[round-mode=places,round-precision=2]{\val}\,%
  \ifdim\fac pt>1pt
    \pos{(\texttimes\num[round-mode=places,round-precision=1]{\fac})}%
  \else
    \ifnum\pdfstrcmp{\fac}{1}=0
      \pos{(\texttimes 1.0)}%
    \else
      \nega{(\texttimes\num[round-mode=places,round-precision=1]{\fac})}%
    \fi
  \fi}%
}

\providecommand{\shadefromtwo}[8]{%
  & \cellcolor{hlrow}#1
  & \cellcolor{hlrow}#2
  & \cellcolor{hlrow}#3
  & \cellcolor{hlrow}#4
  & \cellcolor{hlrow}#5
  & \cellcolor{hlrow}#6
  & \cellcolor{hlrow}#7
  & \cellcolor{hlrow}#8%
}
\newcolumntype{L}[1]{>{\raggedright\arraybackslash}p{#1}}

\begin{table*}[!t]
\centering
\scriptsize
\setlength{\tabcolsep}{3pt}
\renewcommand{\arraystretch}{1.4}

\def\baseVMME{69.22}
\def\baseLVB{65.30}
\def\baseLV{51.39}
\def\baseAVG{61.97}
\def\baseTTFT{4.65}
\resizebox{\textwidth}{!}{%
\begin{tabular}{l
                |l
                |c
                |c
                c
                c
                |c
                |c
                c}
\hline
\makecell{\textbf{Token}\\\textbf{Reduction}} &
\makecell{\textbf{Reduction}\\\textbf{Layers}} &
\makecell{\textbf{\# Red. }\\\textbf{Layers}} &
\makecell{\textbf{VideoMME}\\\textbf{(w/o sub)}\\\textbf{(1$\sim$60m)}} &
\makecell{\textbf{LongVB}\\\textbf{(8s$\sim$60m)}} &
\makecell{\textbf{LVBench}\\\textbf{(30m$\sim$2h)}} &
\makecell{\textbf{Avg}} &
\makecell{\textbf{TTFT (s)}} &
\makecell{\textbf{Reduction} \\\textbf{Overhead (s)}}
\\
\hline

\hline
\rowcolor{black!6}
Baseline & N/A & 0 & 
69.22 & 65.30 & 51.39 & 61.97 & 2.26 &
0 \\
\hline

\multicolumn{9}{l}{\textbf{Test-Time Reduction}} \\
\hline

\multirow{2}{*}{Attn only}
& \makecell{1st Attn}  & 1 & 
\hybvaldelta{68.93}{\baseVMME} &
\hybvaldelta{63.05}{\baseLVB} &
\hybvaldelta{50.23}{\baseLV} &
\hybvaldelta{60.74}{\baseAVG} &
\hybttftfactor{1.12}{\baseTTFT} & 
0.84 \\
& \makecell{All attn} & 6 & 
\hybvaldelta{69.22}{\baseVMME} &
\hybvaldelta{64.62}{\baseLVB} &
\hybvaldelta{51.26}{\baseLV} &
\hybvaldelta{61.70}{\baseAVG} &
\hybttftfactor{1.14}{\baseTTFT} &
0.90 \\
\hline

\multirow{2}{*}{Mamba only}
& \makecell{1st Mamba}  & 1 & 
\hybvaldelta{67.70}{\baseVMME} &
\hybvaldelta{62.60}{\baseLVB} &
\hybvaldelta{48.81}{\baseLV} &
\hybvaldelta{59.70}{\baseAVG} &
\hybttftfactor{1.20}{\baseTTFT} &
0.94 \\
& \makecell{2nd Mamba}  & 1 & 
\hybvaldelta{67.93}{\baseVMME} &
\hybvaldelta{62.23}{\baseLVB} &
\hybvaldelta{50.94}{\baseLV} &
\hybvaldelta{60.37}{\baseAVG} &
\hybttftfactor{1.22}{\baseTTFT} &
0.97 \\
\hline

\multirow{3}{*}{Mamba+Attn}
& \makecell{All attn+1M} & 13 & 
\hybvaldelta{69.22}{\baseVMME} &
\hybvaldelta{63.35}{\baseLVB} &
\textbf{\hybvaldelta{53.07}{\baseLV}} &
\hybvaldelta{61.88}{\baseAVG} &
\hybttftfactor{1.20}{\baseTTFT} &
0.98 \\
& \makecell{All attn+2M} & 20 & 
\textbf{\hybvaldelta{69.26}{\baseVMME}} &
\hybvaldelta{63.05}{\baseLVB} &
\hybvaldelta{52.10}{\baseLV} &
\hybvaldelta{61.47}{\baseAVG} &
\hybttftfactor{1.20}{\baseTTFT} &
0.98 \\
\shadefromtwo{All}{32}
{\hybvaldelta{68.85}{\baseVMME}}{\textbf{\hybvaldelta{64.92}{\baseLVB}}}%
{\hybvaldelta{52.16}{\baseLV}}{\textbf{\hybvaldelta{61.98}{\baseAVG}}}%
{\hybttftfactor{1.21}{\baseTTFT}} 
{1.00} \\
\hline

\multicolumn{9}{l}{\textbf{Train-Time Reduction}} \\
\hline

\multirow{2}{*}{Attn only}
& \makecell{1st attn} & 1 & 
\hybvaldelta{68.74}{\baseVMME} &
\hybvaldelta{64.92}{\baseLVB} &
\hybvaldelta{52.87}{\baseLV} &
\hybvaldelta{62.18}{\baseAVG} &
\hybttftfactor{1.12}{\baseTTFT} &
0.84 \\
\shadefromtwo{All attn}{6}
{\hybvaldelta{69.59}{\baseVMME}}{\textbf{\hybvaldelta{66.12}{\baseLVB}}}%
{\hybvaldelta{54.10}{\baseLV}}{\hybvaldelta{63.27}{\baseAVG}}{\hybttftfactor{1.14}{\baseTTFT}}
{0.90}\\
\hline

\multirow{2}{*}{Mamba only}
& \makecell{1st Mamba} & 1 & 
\hybvaldelta{68.04}{\baseVMME} &
\hybvaldelta{64.10}{\baseLVB} &
\hybvaldelta{51.00}{\baseLV} &
\hybvaldelta{61.05}{\baseAVG} &
\hybttftfactor{1.20}{\baseTTFT} &
0.94 \\
& \makecell{2nd Mamba} & 1 & 
\hybvaldelta{67.07}{\baseVMME} &
\hybvaldelta{63.50}{\baseLVB} &
\hybvaldelta{52.42}{\baseLV} &
\hybvaldelta{61.00}{\baseAVG} &
\hybttftfactor{1.22}{\baseTTFT} & 
0.97 \\
\hline

\multirow{3}{*}{Mamba+Attn}
& \makecell{All attn+1M} & 13 & 
\hybvaldelta{69.00}{\baseVMME} &
\hybvaldelta{63.35}{\baseLVB} &
\hybvaldelta{52.62}{\baseLV} &
\hybvaldelta{61.66}{\baseAVG} &
\hybttftfactor{1.2}{\baseTTFT} & 
0.98\\
& \makecell{All attn+2M} & 20 & 
\hybvaldelta{69.11}{\baseVMME} &
\hybvaldelta{63.80}{\baseLVB} &
\hybvaldelta{52.81}{\baseLV} &
\hybvaldelta{61.91}{\baseAVG} &
\hybttftfactor{1.2}{\baseTTFT} & 
0.98 \\
\shadefromtwo{All}{32}
{\textbf{\hybvaldelta{69.70}{\baseVMME}}}{\hybvaldelta{66.04}{\baseLVB}}%
{\textbf{\hybvaldelta{54.29}{\baseLV}}}{\textbf{\hybvaldelta{63.34}{\baseAVG}}}{\hybttftfactor{1.21}{\baseTTFT}} 
{1.00}\\
\hline
\end{tabular}
}
\caption{\textbf{Token reduction at 25\% compression for Nemotron-Nano-V2 VL 12B with various patterns.} We compare test-time (no finetuning) and train-time reduction (finetuned with reduction). 
"Reduction Layers" indicates where reduction is applied and "\# Red. Layers" counts them; ($\cdot$) denotes change from baseline; TTFT speedups are relative to baseline. 
Patterns: \texttt{1st Attn}/\texttt{1st Mamba}, \texttt{All Attn}, \texttt{All Attn+1M/2M}, and \texttt{All}; see \cref{fig:hybrid_patterns}. 
LLM-stage TTFT (including overhead) and the reduction overheads are measured on a single NVIDIA A100 with 256-frame input.
}
\label{tab:token_reduction_hybrid}
\end{table*}
\subsection{Results}
\cref{tab:token_reduction_hybrid} summarizes reduction schedules on hybrid (Nemotron-Nano-V2 VL 12B) across VideoMME, LongVB, and LVBench with TTFT and reduction overheads. Under test-time reduction (25\% budget), accuracy is largely preserved but depends on where reduction is applied: reducing first attention layer lowers avg. to 60.74 ($-1.23$), while adding Mamba reduction (\texttt{All attn+1M}) recovers long-horizon performance (LVBench 53.07, $+1.68$) and keeps avg. near baseline (61.88, $-0.09$). 

Train-time reduction is consistently stronger: reducing all attention layers (\texttt{All attn}) improves all benchmarks (Avg 63.27, $+1.30$), and reducing all layers (\texttt{All}) performs best (Avg 63.34, $+1.37$). Overall, query-aware reduction improves both efficiency and long-context reasoning.


\paragraph{Comparison with Token-Reduction Baselines.}
\begin{table}[t]
\centering
\scriptsize
\setlength{\tabcolsep}{1.5pt}
\renewcommand{\arraystretch}{1.3}
\setlength{\extrarowheight}{1.2pt}
\resizebox{\linewidth}{!}{%
\begin{tabular}{l c c c c c}
\hline
\textbf{Method} & \textbf{\tiny VideoMME} & \textbf{\tiny LongVB} & \textbf{\tiny  LVBench} & \textbf{\tiny Avg} & \textbf{\tiny Latency} \\
\hline
\rowcolor{black!6}
Baseline 
& 69.22 
& 65.30 
& 51.39 
& 61.93
& 5.57\\

\hline
FastVID~\cite{shen2025fastviddynamicdensitypruning} 
& 69.59 
& 63.65 
& 51.13 
& 61.46
& 3.25\\

CDPruner~\cite{zhang2025attentionsimilaritymaximizingconditional}
& 69.30 
& 53.78 
& 36.86 
& 53.31
& 2.70 \\

\hline
FlexSelect~\cite{zhang2025flexselect} (55\% budget)
& 69.33 
& 65.30 
& 52.61 
& 62.41
& 3.40 \\

FlexSelect-Lite~\cite{zhang2025flexselect} $\dagger$
& 69.22 
& 65.30 
& 51.39 
& 61.97
& 2.07 \\

\hline
\textbf{Ours} 
& 68.85 
& 64.92 
& 52.16 
& 61.98
& 2.13 \\

\textbf{Ours $\dagger$} 
& 69.70 
& \textbf{66.04} 
& 54.29 
& 63.34
& 2.13 \\
\rowcolor{blue!6}
\textbf{FastVID~\cite{shen2025fastviddynamicdensitypruning}  + Ours} 
& \textbf{70.41} 
& 65.97 
& \textbf{54.74} 
& \textbf{63.71}
& 3.25 \\


\hline
\end{tabular}
}
\caption{\textbf{Comparison with token-reduction baselines on the same Nemotron-Nano-V2 VL 12B backbone.} All methods use a 25\% token budget, except FlexSelect (55\%). $\dagger$ denotes trained variants. Ours is the strongest standalone at 25\%, and combining with FastVID yields further gains, highlighting complementarity between redundancy and query-conditioned reduction.}
\label{tab:baseline_comparison_hybrid12b}
\vspace{-0.2cm}
\end{table}
\cref{tab:baseline_comparison_hybrid12b} compares our method with prior token-reduction approaches on Nemotron-Nano-V2 VL 12B, including CDPruner~\cite{zhang2025attentionsimilaritymaximizingconditional}, a training-free query-conditioned method that selects visually diverse and instruction-relevant tokens, FastVID~\cite{shen2025fastviddynamicdensitypruning}, a query-agnostic method that reduces visual redundancy through temporal segmentation and density-based spatiotemporal pruning, and FlexSelect~\cite{zhang2025flexselect} that performs query-conditioned tokens selection from a reference layer (or via a learned selector in FlexSelect-Lite). 
CDPruner degrades significantly in our hybrid long-video setting, especially on LVBench, while FastVID is more competitive but remains below our hybrid-aware method. 

In FlexSelect, following the original setup, we use layer 25, which limits compression to 55\% since earlier layers process all tokens. Even at this higher budget, FlexSelect underperforms our train-time result at 25\%. FlexSelect-Lite reaches 25\% compression but only matches the baseline, while our train-time method exceeds it by +1.37.
Our train-time variant achieves the best overall result at 25\%. 

Importantly, our method is complementary to query-agnostic token-reduction approaches, which remove visual redundancy independently of the query, while ours performs query-conditioned selection within the backbone. For example, combining our method with FastVID yields further gains over either alone, achieving the best overall performance (63.71) without training. This suggests that our method can be effectively composed with other redundancy reduction methods.

\definecolor{posblue}{RGB}{0,60,200}
\definecolor{negred}{RGB}{200,0,0}


\providecommand{\valdelta}[2]{%
  \edef\dlt{\fpeval{#1-(#2)}}%
  \num{#1}\ %
  \ifdim\dlt pt=0pt
    \pos{(+\num{0})}%
  \else
    \ifdim\dlt pt>0pt
      \pos{(+\num{\dlt})}%
    \else
      \nega{(-\num{\fpeval{-(\dlt)}})}%
    \fi
  \fi
}

\providecommand{\ttftfactor}[2]{%
  \edef\val{\fpeval{round(#1,2)}}%
  \edef\fac{\fpeval{round((#2)/(#1),1)}}%
  \val\ %
  \ifdim\fac pt>1pt
    \pos{(\texttimes\fac)}%
  \else
    \ifnum\pdfstrcmp{\fac}{1}=0
      \pos{(\texttimes 1.00)}%
    \else
      \nega{(\texttimes\fac)}%
    \fi
  \fi
}

\def\baseTTFThybrid{4.65}
\def\baseTTFTqwen{4.78}


\begin{table*}[htbp]
\centering
\scriptsize
\setlength{\tabcolsep}{1pt}
\renewcommand{\arraystretch}{1.3}
\setlength{\extrarowheight}{1.2pt}

\def\nemoVMME{69.22}
\def\nemoLVB{65.30}
\def\nemoLV{51.39}
\def\nemoAVG{61.97}

\def\qwenVMME{69.89}
\def\qwenLVB{66.64}
\def\qwenLV{50.94}
\def\qwenAVG{62.49}

\resizebox{\textwidth}{!}{%
\begin{tabular}{C{1.5cm}
                |C{1.5cm}
                |C{1.9cm}
                C{1.9cm}
                C{1.9cm}
                |C{1.7cm}
                |C{1.7cm}}
\hline
\textbf{\tiny Token Reduction} &
\makecell{\textbf{\tiny Comp.}\\\textbf{\tiny Rate (\%)}} &
\makecell{\textbf{\tiny VideoMME}\\\textbf{\tiny (1$\sim$60m)}} &
\makecell{\textbf{\tiny LongVB}\\\textbf{\tiny (8s$\sim$60m)}} &
\makecell{\textbf{\tiny LVBench}\\\textbf{\tiny (30m$\sim$2h)}} &
\textbf{\tiny Avg} & \textbf{\tiny TTFT(s)}\\
\hline

\multicolumn{7}{c}{\textbf{Nemotron-Nano-V2 VLM 12B}} \\
\hline
\rowcolor{black!6}
\makecell{Baseline} & 100 &
69.22 & 65.30 & 51.39 & 61.97 & 4.65 \\
\hline

\multirow{3}{=}{\makecell{All\\attention\\layers}} &
50.1 &
\valdelta{70.00}{\nemoVMME} &
 \valdelta{65.74}{\nemoLVB} &
\valdelta{54.42}{\nemoLV} &
 \valdelta{63.39}{\nemoAVG}  &
 \ttftfactor{2.21}{\baseTTFThybrid}\\
\cline{2-7}

& 34.7 &
\valdelta{70.26}{\nemoVMME} &
 \valdelta{66.04}{\nemoLVB} &
 \valdelta{54.29}{\nemoLV} &
 \valdelta{63.53}{\nemoAVG}  &
\ttftfactor{1.53}{\baseTTFThybrid}\\
\cline{2-7}

& 25.2 &
\valdelta{69.59}{\nemoVMME} &
\valdelta{66.12}{\nemoLVB} &
\valdelta{54.10}{\nemoLV} &
\valdelta{63.27}{\nemoAVG} &
\ttftfactor{1.14}{\baseTTFThybrid}\\
\hline
\hline

\multicolumn{7}{c}{\textbf{Qwen3-VL 8B}} \\
\hline
\rowcolor{black!6}
Baseline & 100 & 69.89 & 66.64 & 50.94 & 62.49 & 4.78 \\
\hline

\multirow{3}{=}{\makecell{All\\attention\\layers}} &
50.1 &
\valdelta{70.52}{\qwenVMME} &
\valdelta{66.79}{\qwenLVB} &
\valdelta{46.93}{\qwenLV} &
\valdelta{61.41}{\qwenAVG} &
\ttftfactor{2.37}{\baseTTFTqwen}\\
\cline{2-7}

& 34.7 &
\valdelta{68.93}{\qwenVMME} &
\valdelta{66.12}{\qwenLVB} &
\valdelta{45.13}{\qwenLV} &
\valdelta{60.06}{\qwenAVG} &
\ttftfactor{1.61}{\baseTTFTqwen}\\
\cline{2-7}

& 25.1 &
\valdelta{68.41}{\qwenVMME} &
\valdelta{64.62}{\qwenLVB} &
\valdelta{43.19}{\qwenLV} &
\valdelta{58.74}{\qwenAVG} &
\ttftfactor{1.18}{\baseTTFTqwen}\\
\hline
\end{tabular}
}

\caption{
\textbf{Varying compression rates across architectures.} We apply trained reduction at all attention layers (6 layers) for Nemotron-Nano-V2 VL and Qwen3-VL. “Comp. Rate” is the remaining token ratio; (·) denotes change from baseline. TTFT is measured on an NVIDIA A100 at 256 frames.
}
\vspace{-0.5cm}
\label{tab:compression_rate}
\end{table*}

\paragraph{Compression Effects Across Architectures.}
\cref{tab:compression_rate} varies token budgets (50\%–25\%) for hybrid and Transformer models with fixed six-layer attention reduction and brief finetuning. The architectures show distinct speed–quality trade-offs. The hybrid model (Nemotron-Nano-V2) improves performance across all budgets (Avg $+1.3$–$+1.56$) while reducing TTFT from 4.65\,s to 2.21–1.14\,s (2.1–4.1$\times$). In contrast, the Transformer (Qwen3-VL) improves only at 50\% and degrades at lower budgets. This supports our analysis (\cref{sec:analysis}) that early compression in Transformers causes unrecoverable information loss. Trends are consistent across schedules (\cref{subsec:additional_results}). Overall, reduction improves both speed and quality in hybrids, while Transformers exhibit a standard trade-off.

\begin{figure*}[t]
  \centering
  \begin{subfigure}[b]{0.50\textwidth}
  \centering
   \resizebox{\linewidth}{!}{%
   \begin{tikzpicture}
\begin{axis}[
  width=2.1in,
  height=1.40in,
  xmode=log,
  log basis x=2,
  xmin=14,
  xmax=292,
  xtick={16,32,64,128,256},
  xticklabels={16,32,64,128,256},
  xlabel={Frames},
  ylabel={Latency (s)},
  xlabel style={font=\scriptsize, yshift=2pt},
  ylabel style={font=\scriptsize, yshift=-3pt},
  ymin=0,
  ymax=5.8,
  ytick={0,1,2,3,4,5},
  tick label style={font=\scriptsize},
  label style={font=\scriptsize},
  legend style={
    font=\tiny,
    draw=none,
    fill=none,
    cells={anchor=west},
    at={(0.03,0.97)},
    anchor=north west,
    row sep=-1pt,
    inner sep=1pt
  },
  grid=major,
  grid style={gray!18},
  every axis plot/.append style={
    line width=0.9pt,
    mark size=1.15pt
  },
]
\addplot[color=black, mark=*] coordinates {
  (16,0.349) (32,0.677) (64,1.313) (128,2.714) (256,5.567)
};
\addplot[color=blue!70!black, mark=square*] coordinates {
  (16,0.179) (32,0.287) (64,0.520) (128,1.033) (256,2.052)
};
\addplot[color=orange!85!black, mark=triangle*] coordinates {
  (16,0.214) (32,0.327) (64,0.564) (128,1.080) (256,2.125)
};
\legend{No reduction,Attn only 25\%,Attn+Mamba 25\%}
\end{axis}
\end{tikzpicture}


   }
  \caption{{{End-to-end inference latency}}}
  \label{fig:end_to_end_latency_hybrid}
  \end{subfigure}
  \hspace{0.02\textwidth}
  \begin{subfigure}[b]{0.42\textwidth}
  \centering
  \includegraphics[width=1.0\linewidth]{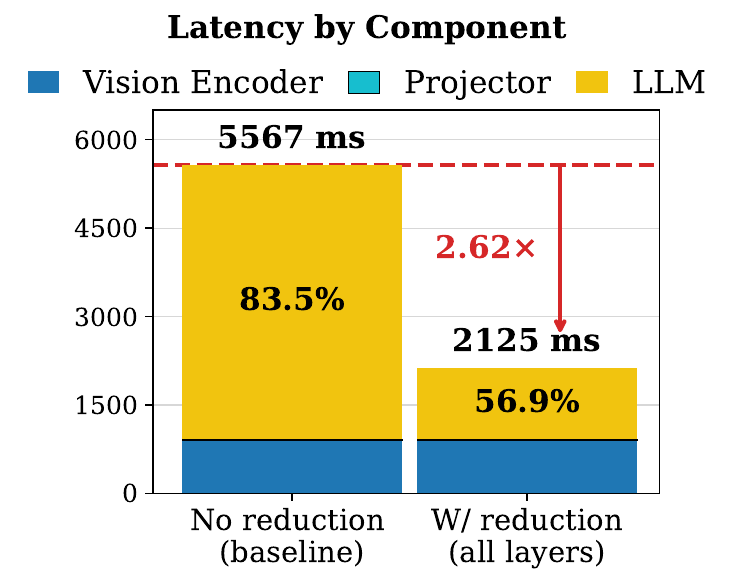}
  \caption{Component-wise latency }
  \label{fig:latency_component_hybrid}
\end{subfigure}
\caption{\textbf{Inference-latency gains from token reduction for Nemotron-Nano-V2 VL 12B.} \textbf{(i)} Measured on a single A100 with 256 frames; token reduction lowers end-to-end latency, achieving up to 2.71$\times$ speedup. \textbf{(ii)} Vision encoder, projector, and LLM latency for baseline vs.\ all-layer reduction on 256-frame input.}
\end{figure*}

\subsection{Efficiency Analysis}
\paragraph{Latency Analysis.}
\cref{tab:token_reduction_hybrid} reports TTFT and reduction overhead for Nemotron-Nano-V2 on an A100 with 256-frame input. At $\sim$25\% token budget, TTFT drops from 4.65\,s to 1.12--1.22\,s with 0.84--1.00\,s overhead, achieving 3.8–4.2$\times$ speedups. Adding Mamba reduction slightly increases overhead but maintains speedups. 
Including vision encoder and projector costs, \cref{fig:end_to_end_latency_hybrid} shows end-to-end latency decreases from 5.567\,s to 2.125\,s (2.62$\times$) at 256 frames, with the gap widening for longer videos as LLM cost scales faster. Without reduction, latency grows steeply and reaches out-of-memory (OOM) at 512 frames (see \cref{fig:latency_hybrid} in Appendix). 
\cref{fig:latency_component_hybrid} further decomposes latency into vision encoder, projector, and LLM stages, showing the baseline is dominated by the LLM (83.5\%). With all-layer reduction, LLM latency drops to 56.9\%, confirming that long-video inference is primarily bounded by LLM compute.



\paragraph{Training Efficiency.}
Token reduction also improves training efficiency. As shown in \cref{tab:training_efficiency}, with 64-frame inputs, 25\% reduction yields 1.48$\times$ faster training, 1.45$\times$ higher throughput, and 8.7\% lower peak GPU memory. More importantly, under the same A100 80GB memory budget, token reduction supports 128-frame finetuning, whereas the no-reduction baseline is limited to 96 frames. Thus, token reduction not only accelerates training but also makes longer-video training more feasible under fixed hardware constraints.

\subsection{Ablations}
\definecolor{posblue}{RGB}{0,60,200}
\definecolor{negred}{RGB}{200,0,0}

\newcommand{\scorevaldelta}[2]{%
  \edef\dlt{\fpeval{#1-(#2)}}%
  \makecell[c]{\num{#1}\\[-0.15em]{\tiny
  \ifdim\dlt pt=0pt
    \pos{(+\num{0.00})}%
  \else
    \ifdim\dlt pt>0pt
      \pos{(+\num[minimum-decimal-digits=2]{\dlt})}%
    \else
      \nega{(-\num[minimum-decimal-digits=2]{\fpeval{-(\dlt)}})}%
    \fi
  \fi
  }}%
}

\newcommand{\scorevaldeltaB}[2]{%
  \edef\dlt{\fpeval{#1-(#2)}}%
  \makecell[c]{\textbf{\num{#1}}\\[-0.15em]{\tiny
  \ifdim\dlt pt=0pt
    \pos{(+\num{0.00})}%
  \else
    \ifdim\dlt pt>0pt
      \pos{(+\num[minimum-decimal-digits=2]{\dlt})}%
    \else
      \nega{(-\num[minimum-decimal-digits=2]{\fpeval{-(\dlt)}})}%
    \fi
  \fi
  }}%
}

\begin{table}[H]
\centering
\footnotesize
\setlength{\tabcolsep}{2.4pt}
\renewcommand{\arraystretch}{1.24}

\def\baseVMME{69.22}
\def\baseLVB{65.30}
\def\baseLV{51.39}
\def\baseAVG{61.97}

\resizebox{\linewidth}{!}{%
\begin{tabular}{@{}C{1.15cm}|C{1.62cm}|C{1.00cm}C{1.30cm}C{1.40cm}|C{0.95cm}@{}}
\hline
\makecell{\textbf{Pattern}} &
\makecell{\textbf{Score}} &
\makecell{\textbf{Video}\\\textbf{MME}} &
\makecell{\textbf{LongVB}} &
\makecell{\textbf{LVBench}} &
\makecell{\textbf{Avg}} \\
\hline
\rowcolor{black!6}
Baseline & N/A &
69.22 & 65.30 & 51.39 & 61.97 \\
\hline

\multirow{2}{*}{\makecell{All attn\\+ 1M}}
& \makecell{Full\\Dynamic}
& \scorevaldelta{60.93}{\baseVMME}
& \scorevaldelta{57.67}{\baseLVB}
& \scorevaldelta{44.16}{\baseLV}
& \scorevaldelta{54.25}{\baseAVG} \\
\cline{2-6}
& \makecell{\textbf{Decoupled}\\\textbf{(Ours)}}
& \scorevaldeltaB{69.22}{\baseVMME}
& \scorevaldeltaB{63.35}{\baseLVB}
& \scorevaldeltaB{53.07}{\baseLV}
& \scorevaldeltaB{61.88}{\baseAVG} \\
\hline

\multirow{2}{*}{\makecell{All attn\\+ 2M}}
& \makecell{Full\\Dynamic}
& \scorevaldelta{60.63}{\baseVMME}
& \scorevaldelta{57.44}{\baseLVB}
& \scorevaldelta{43.58}{\baseLV}
& \scorevaldelta{53.88}{\baseAVG} \\
\cline{2-6}
& \makecell{\textbf{Decoupled}\\\textbf{(Ours)}}
& \scorevaldeltaB{69.26}{\baseVMME}
& \scorevaldeltaB{63.05}{\baseLVB}
& \scorevaldeltaB{52.10}{\baseLV}
& \scorevaldeltaB{61.47}{\baseAVG} \\
\hline

\multirow{2}{*}{\makecell{All\\layers}}
& \makecell{Full\\Dynamic}
& \scorevaldelta{68.37}{\baseVMME}
& \scorevaldelta{62.98}{\baseLVB}
& \scorevaldelta{50.42}{\baseLV}
& \scorevaldelta{60.59}{\baseAVG} \\
\cline{2-6}
& \makecell{\textbf{Decoupled}\\\textbf{(Ours)}}
& \scorevaldeltaB{68.85}{\baseVMME}
& \scorevaldeltaB{64.92}{\baseLVB}
& \scorevaldeltaB{52.16}{\baseLV}
& \scorevaldeltaB{61.98}{\baseAVG} \\
\hline
\end{tabular}
}

\caption{\textbf{Importance Score Ablation for Mamba Layers.} Training-free reduction at $\sim$25\% token budget.
We compare \emph{Full Mamba Dynamic} (Eq.~\ref{eq:ssm_weight}, decay) vs.\ \emph{Decoupled Content Alignment} (Eq.~\ref{eq:ssm_importance}, no decay). 
Full Mamba Dynamic degrades performance by up to $8.09$ points when Mamba layers are used, as cumulative decay $\prod_u \bar{\mA}_u$ suppresses early visual tokens regardless of relevance. Removing decay preserves near-baseline accuracy and improves LVBench. ($\cdot$) denotes change vs.\ baseline.
}
\label{tab:scoring_ablation}
\end{table}
\paragraph{Importance Score Design: Decay Term Ablation.} 
A central methodological claim of this paper is that the \emph{Full Mamba Dynamic} (\cref{eq:ssm_weight}) is a poor importance signal compared to \emph{Decoupled Content Alignment} (\cref{eq:ssm_importance}). \cref{tab:scoring_ablation} validates this claim directly. Under identical training-free reduction at $\sim$25\%, using the full dynamic drops Avg by $7.72$ (\texttt{All attn+1M}) and $8.09$ (\texttt{All attn+2M}), with large LVBench degradations ($-7.23$, $-7.81$). When applied at all layers, the gap shrinks to $1.38$ as decay weakens in deeper layers. In all cases, the decoupled score matches or exceeds the baseline, reflecting intrinsic scoring quality independent of finetuning. These results suggest that cumulative decay $\prod_u \bar{\mA}_u$ should be excluded for token selection, leaving alignment $\bar{\vb}_j^\top \vc_t$ as the key signal, a principle that applies to other SSM/linear-attention hybrids.

\paragraph{Mamba Token-Reduction Patterns.}
\cref{tab:ablation} (Left) studies how to insert Mamba-layer reduction when applied to all attention layers (\cref{fig:hybrid_patterns}). Adding reduction at the second inter-attention position (\texttt{A+1M(L2)}) matches baseline (69.22), while two middle reductions (\texttt{A+2M(L2,3)}) slightly improve it (69.26, $+0.04$). In contrast, including the earliest position (\texttt{A+2M(L1,3)}) degrades performance (68.26, $-0.96$), indicating early reduction harms later processing, consistent with \cref{sec:analysis}.


\paragraph{Reduction Types.}
\cref{tab:ablation} (Right) compares query-based token selection and average pooling. Query-based selection achieves the best accuracy (69.70, +0.48) with a 1.9$\times$ speedup. In contrast, replacing Mamba reduction with average pooling (\texttt{Query (T) + AvgPool (M)}) lowers VideoMME to 66.15 ($-3.07$), and pooling everywhere performs worst (65.67, $-3.55$) despite similar speedups. Overall, query-based selection better preserves task-relevant information under heavy compression.


\definecolor{posblue}{RGB}{0,60,200}
\definecolor{negred}{RGB}{200,0,0}


\providecommand{\valdelta}[2]{%
  \edef\dlt{\fpeval{#1-(#2)}}%
  \num{#1}\ %
  \ifdim\dlt pt=0pt
    \pos{(+\num{0})}%
  \else
    \ifdim\dlt pt>0pt
      \pos{(+\num{\dlt})}%
    \else
      \nega{(-\num{\fpeval{-(\dlt)}})}%
    \fi
  \fi
}

\providecommand{\ttftfactor}[2]{%
  \edef\val{\fpeval{round(#1,2)}}%
  \edef\fac{\fpeval{round((#2)/(#1),1)}}%
  \val\ %
  \ifdim\fac pt>1pt
    \pos{(\texttimes\fac)}%
  \else
    \ifnum\pdfstrcmp{\fac}{1}=0
      \pos{(\texttimes 1.00)}%
    \else
      \nega{(\texttimes\fac)}%
    \fi
  \fi
}

\def\baseTTFT{2.26}


\definecolor{posblue}{RGB}{0,60,200}
\definecolor{negred}{RGB}{200,0,0}


\providecommand{\valdelta}[2]{%
  \edef\dlt{\fpeval{#1-(#2)}}%
  \num{#1}\ %
  \ifdim\dlt pt=0pt
    \pos{(+\num{0})}%
  \else
    \ifdim\dlt pt>0pt
      \pos{(+\num{\dlt})}%
    \else
      \nega{(-\num{\fpeval{-(\dlt)}})}%
    \fi
  \fi
}

\providecommand{\ttftfactor}[2]{%
  \edef\val{\fpeval{round(#1,2)}}%
  \edef\fac{\fpeval{round((#2)/(#1),1)}}%
  \val\ %
  \ifdim\fac pt>1pt
    \pos{(\texttimes\fac)}%
  \else
    \ifnum\pdfstrcmp{\fac}{1}=0
      \pos{(\texttimes 1.00)}%
    \else
      \nega{(\texttimes\fac)}%
    \fi
  \fi
}

\def\baseTTFT{2.25}


\begin{table*}[!t]
\centering

\begin{minipage}[t]{0.42\textwidth}
\vspace{0pt}
\centering
\scriptsize
\setlength{\tabcolsep}{3pt}
\renewcommand{\arraystretch}{1.3}
\setlength{\extrarowheight}{1.2pt}

\def\nemoVMME{69.22}
\def\nemoLVB{65.30}
\def\nemoLV{51.39}
\def\nemoAVG{61.97}

\def\qwenVMME{69.89}
\def\qwenLVB{66.64}
\def\qwenLV{50.94}
\def\qwenAVG{62.49}

\begin{tabular}{C{3.0cm}
                |C{2.8cm}}
\hline
\makecell{\textbf{Reduction Pattern}} &
\makecell{\textbf{Video-MME} $\uparrow$} \\
\hline

\rowcolor{black!6}
\makecell{Baseline} & 69.22 \\
\hline

{A + 1M (L2)} &
\valdelta{69.22}{\nemoVMME} \\

A + 2M (L2,3) &
\valdelta{69.26}{\nemoVMME} \\

A + 2M (L1,3) &
\valdelta{68.26}{\nemoVMME} \\
\hline

\end{tabular}
\end{minipage}
\hfill
\begin{minipage}[t]{0.56\textwidth}
\vspace{0pt}
\centering
\scriptsize
\setlength{\tabcolsep}{3pt}
\renewcommand{\arraystretch}{1.3}
\setlength{\extrarowheight}{1.2pt}

\def\nemoVMME{69.22}
\def\nemoLVB{65.30}
\def\nemoLV{51.39}
\def\nemoAVG{61.97}

\def\qwenVMME{69.89}
\def\qwenLVB{66.64}
\def\qwenLV{50.94}
\def\qwenAVG{62.49}

\begin{tabular}{C{4.0cm}
                |C{2.1cm}
                |C{1.8cm}}
\hline
\makecell{\textbf{Reduction Type}} &
\makecell{\textbf{Video-MME} $\uparrow$} &
\makecell{\textbf{TTFT $\downarrow$ (s)}}\\
\hline

\rowcolor{black!6}
\makecell{Baseline} & 69.22 &
4.65 \\
\hline

{Query-based only} &
\valdelta{69.70}{\nemoVMME} &
\ttftfactor{1.21}{\baseTTFT}\\

Query (T) + AvgPool (M) &
\valdelta{66.15}{\nemoVMME} &
\ttftfactor{1.18}{\baseTTFT}\\

AvgPool only &
\valdelta{65.67}{\nemoVMME}  &
\ttftfactor{1.16}{\baseTTFT}\\
\hline

\end{tabular}
\end{minipage}

\caption{\textbf{Mamba Token Reduction Patterns and Reduction Types at 25\% budget}. (left) Reduction is applied to all attention (A) and selected Mamba (M) layers. 
(L$\cdot$) denotes Mamba layer indices where reduction is applied (see \cref{fig:hybrid_patterns}). 
(right) Query-based reduction (A) is compared with average pooling (M). 
TTFT measures LLM-stage latency on a single A100 with 256-frame input.}
\label{tab:ablation}
\end{table*}
\begin{table}[t]
\centering
\scriptsize
\setlength{\tabcolsep}{2pt}
\renewcommand{\arraystretch}{1.15}
\resizebox{\columnwidth}{!}{%
\begin{tabular}{lccc}
\toprule
\textbf{Metric} & \makecell{\textbf{No}\\\textbf{Reduction}} & \makecell{\textbf{25\%}\\\textbf{Reduction}} & \textbf{Gain} \\
\midrule
Train runtime & 1251.7s & 862.3s & 1.45$\times$ \\
Wall clock & 1390s & 942s & 1.48$\times$ \\
Samples/sec & 0.818 & 1.187 & 1.45$\times$ \\
Peak GPU memory & 77,436 MiB & 70,664 MiB & 8.7\% saved \\
Max frames/sample & 96 & 128 & +33\% \\
\bottomrule
\end{tabular}
}
\caption{\textbf{Training efficiency with token reduction.} Runtime metrics are measured under the same training setup with 64-frame inputs. The maximum frames per sample are measured under the same A100 80GB memory budget.}
\label{tab:training_efficiency}
\end{table}


\subsection{Additional Results}
\label{subsec:additional_results}
\paragraph{Transformer Results.}

\cref{tab:token_reduction_transformer} (Appendix) reports train-time token reduction on Qwen3-VL 8B using the same positions as hybrids. While TTFT improves significantly (4.78\,s $\rightarrow$ 1.07--1.18\,s, up to 4.5$\times$), accuracy drops more than in hybrids, especially on LVBench, with Avg decreasing by $\sim$3--4 points. Early-layer reduction is particularly harmful, consistent with \cref{sec:analysis}, as stateless reduction in Transformers causes information loss. \cref{fig:latency_transformer} (Appendix) shows similar latency trends, with up to 2.36$\times$ end-to-end speedup at 256 frames.



\paragraph{Comparisons with Other VLMs.}
\cref{tab:comparison} (Appendix) compares our method with proprietary models, open-source video LLMs, and prior token-selection approaches. Direct parameter-matched comparison is not always possible: no public \emph{hybrid} video VLM exists in the 8B range, and existing 8B baselines differ in vision backbones and training. Our goal is to show that our architecture-aware reduction is effective for hybrids and competitive with strong Transformer methods. Despite these differences, our approach narrows the gap to much larger models (e.g., 72B) on LongVideoBench and LVBench while remaining competitive on VideoMME.




\section{Related Works}
\label{sec:related}

Token reduction is widely used in multimodal LLMs to reduce redundant computation, especially for long-video inputs. Prior work includes query-aware selection based on language–vision relevance~\cite{chen2024imageworth12tokens,zhang2025sparsevlmvisualtokensparsification}, token merging~\cite{hyun2025multi}, and long-video compression methods such as FlexSelect, DynTok, and METok~\cite{zhang2025flexselect,zhang2025dyntok,wang2025metok}. These approaches are largely developed for Transformer architectures and operate primarily on attention layers. Reduction strategies also vary from pruning from early layers~\cite{chen2024imageworth12tokens,zhu2025visionselector} to progressive schemes across layers~\cite{xing2024pyramiddrop,liu2025laco,yang2025visionzip}. Our work targets Mamba--Transformer hybrid VLMs and explicitly models their stateful information flow, using depth-dependent importance and recurrent state accumulation. A detailed review is provided in ~\Cref{sec:related_appendix} (Appendix).

\section{Conclusion and Limitations}
\label{sec:conclusion}

We study token reduction for long-video VLMs with a focus on Mamba--Transformer hybrids. We show that token importance is unstable across layers and that reduction is \emph{stateful}, allowing information from discarded tokens to persist. We propose a progressive reduction schedule and a unified query-conditioned scoring method for attention and Mamba layers. Our approach achieves substantial speedups at aggressive budgets while maintaining near-baseline accuracy, with gains from light finetuning.

There are several limitations. The reduction schedule is fixed and not input-adaptive or learnable, and our evaluation focuses on visual token reduction in long-video benchmarks, leaving other modalities and deployment scenarios for future work. While we provide strong empirical evidence for stateful compression in hybrid models, a formal theoretical characterization remains an open direction.

\section{Acknowledgments}
We would like to thank the NVIDIA Nemotron team for their help on the hybrid model and the NVIDIA VILA team for their support on the multimodal LLM codebase. We also thank Orazio Gallo, Abhishek Badki, and Hang Su for insightful discussions.
\clearpage

{
  \small
  \bibliographystyle{unsrt}
  \bibliography{bib}
}

\appendix
\clearpage
\appendix

\section{Appendix}

\subsection{The Implicit Attention Mechanism of Mamba}
\label{sec:appendix_ssm_derivation}

We derive the implicit attention structure in selective state-space models (Mamba), showing how each output token aggregates information from all preceding tokens through learned, input-dependent weights.

\subsubsection{State-Space Recurrence}
Consider the discrete-time state-space recurrence in Mamba-2~\citep{dao2024transformersssmsgeneralizedmodels}:
\begin{align}
\mS_t &= \bar{\mA}_t \mS_{t-1} + \vx_t (\Delta_t \vb_t)^\top, \label{eq:ssm_state_update} \\
\vy_t &= \mS_t \vc_t, \label{eq:ssm_output}
\end{align}
where $\mS_t \in \mathbb{R}^{d \times n}$ is the hidden state matrix, $\bar{\mA}_t = \exp(-\Delta_t \mA)$ is the discretized state transition (a diagonal matrix with entries in $(0,1)$), $\vx_t \in \mathbb{R}^d$ is the input, $\vb_t, \vc_t \in \mathbb{R}^n$ are input-dependent projection vectors, and $\Delta_t > 0$ is the discretization step size.

\subsubsection{Unrolling the Recurrence}
Substituting \cref{eq:ssm_state_update} into itself recursively:
\begin{align}
\mS_t &= \bar{\mA}_t \mS_{t-1} + \vx_t (\Delta_t \vb_t)^\top \nonumber \\
&= \bar{\mA}_t \left( \bar{\mA}_{t-1} \mS_{t-2} + \vx_{t-1} (\Delta_{t-1} \vb_{t-1})^\top \right) \nonumber \\
&\quad + \vx_t (\Delta_t \vb_t)^\top \nonumber \\
&= \sum_{j=1}^{t} \left( \prod_{u=j+1}^{t} \bar{\mA}_u \right) \vx_j (\Delta_j \vb_j)^\top. \label{eq:ssm_unrolled}
\end{align}
Substituting into \cref{eq:ssm_output}:
\begin{align}
\vy_t &= \mS_t \vc_t = \sum_{j=1}^{t} \left( \prod_{u=j+1}^{t} \bar{\mA}_u \right) \vx_j (\Delta_j \vb_j)^\top \vc_t \nonumber \\
&= \sum_{j=1}^{t} \underbrace{\left( \prod_{u=j+1}^{t} \bar{\mA}_u \right) (\Delta_j \vb_j)^\top \vc_t}_{w_{t,j}} \vx_j. \label{eq:ssm_attention_form}
\end{align}

This reveals an attention-like structure: the output $\vy_t$ is a weighted combination of all past inputs $\{\vx_j\}_{j=1}^t$. The scalar weights $w_{t,j}$ depend on three factors:
\begin{enumerate}
    \item \textbf{Content alignment}: $\vb_j^\top \vc_t$ measures the relevance between the input at position $j$ (encoded by $\vb_j$) and the query at position $t$ (encoded by $\vc_t$).
    \item \textbf{Input gating}: $\Delta_j$ controls how strongly position $j$ writes to the state.
    \item \textbf{Temporal decay}: $\prod_{u=j+1}^{t} \bar{\mA}_u$ exponentially decays contributions from distant positions.
\end{enumerate}

\cref{eq:ssm_attention_form} suggests an intuitive role for $\vb$ and $\vc$, we next formalize this by mapping the recurrence directly to the key-query formulation of Linear Attention. This connection justifies our use of these projections for token selection.

\subsection{Token Selection via Mamba's Recurrent Updates}
\label{sec:appendix_linear_attention_connection}

Following existing works in analyzing the associative memory mechanism in attention~\citep{katharopoulos2020transformers,deltanet,deltaformer}, we draw a connection between standard Softmax attention and the recurrence in state-space models. This elucidates why the projections $\vb$ and $\vc$ function as keys and queries, respectively, and justifies their use as an importance metric.

\subsubsection{Attention via Kernel Functions}
We begin by expressing softmax attention in a more general form using kernel functions~\citep{deltaformer}, which will reveal its structural similarity to Mamba's recurrence. Standard softmax attention computes the output as a weighted sum of values:
\begin{equation}
\vy_t = \sum_{j=1}^{t} \frac{\exp(\vq_t^\top \vk_j / \sqrt{d})}{\sum_{i=1}^{t} \exp(\vq_t^\top \vk_i / \sqrt{d})} \vv_j,
\end{equation}
The key observation is that the exponential term $\exp(\vq_t^\top \vk_j / \sqrt{d})$ can be viewed as a \emph{kernel function} $\kappa(\vq_t, \vk_j)$ that measures similarity between the query and key~\citep{katharopoulos2020transformers}. A kernel function can always be decomposed as an inner product in some (possibly high-dimensional) feature space: $\kappa(\vq, \vk) = \phi(\vq)^\top \phi(\vk)$, where $\phi(\cdot)$ is a feature map. This decomposition allows us to express the attention weight as a normalized inner product in feature space. By defining the normalization factor as $Z_t = \sum_{i=1}^{t} \phi(\vq_t)^\top \phi(\vk_i)$, attention can be written as:

\begin{equation}
\label{eq:attention_kernel_form}
\vy_t = \sum_{j=1}^{t} \underbrace{\frac{1}{Z_t}}_{\text{Normalization}} \underbrace{\left( \phi(\vq_t)^\top \phi(\vk_j) \right)}_{\text{Attention Score}} \vv_j.
\end{equation}


This reformulation reveals that attention computes a weighted sum of values, where each weight is determined by the similarity between the query and key in a feature space induced by $\phi$. Different choices of the feature map $\phi$ yield different attention mechanisms: standard softmax attention corresponds to an exponential feature map $\phi(\vx) = \exp(\vx / \sqrt{d})$, while \emph{linear attention}~\citep{katharopoulos2020transformers} uses the identity map $\phi(\vx) = \vx$, giving $\kappa(\vq, \vk) = \vq^\top \vk$. The advantage of this kernel view is that when $\phi$ is simple (e.g., identity), the computation can be reorganized to avoid the quadratic complexity of standard attention.

As we show below, Mamba's recurrence can be viewed through this same lens, where the projections $\vb$ and $\vc$ play the roles of keys and queries, respectively. In this kernel form, the output decomposes into two factors: a normalization term (global $1/Z_t$ for attention, or temporal decay for Mamba) and a content-based alignment score $\phi(\vq_t)^\top \phi(\vk_j)$ that measures query-key compatibility.

\subsubsection{The Implicit Attention in Mamba}
Comparing this to the unrolled Mamba-2 output derived in \cref{eq:ssm_attention_form}, we define the cumulative decay term $\alpha_{t,j}$ to represent the aggregate state transition from step $j$ to $t$: $\alpha_{t,j} = \prod_{u=j+1}^{t} \bar{\mA}_u$. The Mamba update rule can then be written in a form structurally identical to \cref{eq:attention_kernel_form}:
\begin{equation}
\vy_t = \sum_{j=1}^{t} \underbrace{\alpha_{t,j}}_{\text{Decay}} \underbrace{\left( \vc_t^\top (\Delta_j \vb_j) \right)}_{\text{Attention Score}} \vx_j.
\label{eq:mamba_implicit_attention}
\end{equation}

This reveals a direct correspondence between the components of Softmax attention and Mamba:
\begin{itemize}
    \item \textbf{Query Mapping}: The term $\vc_t$ corresponds exactly to the mapped query $\phi(\vq_t)$.
    \item \textbf{Key Mapping}: The term $\Delta_j \vb_j$ corresponds exactly to the mapped key $\phi(\vk_j)$.
    \item \textbf{Alignment Score}: The dot product $\vc_t^\top (\Delta_j \vb_j)$ acts as the kernel similarity $\kappa(\vq_t, \vk_j)$, measuring the compatibility between position $t$ and position $j$.
\end{itemize}

The primary difference lies in the weighting term: whereas Attention uses a global normalization $1/Z_t$ (ensuring weights sum to 1), Mamba uses a time-dependent decay $\alpha_{t,j}$ (ensuring older context fades away).

\subsubsection{Resulting Token Selection Metric}
\label{sec:appendix_token_selection}

\begin{figure}[t]
  \centering
  \includegraphics[width=0.6\linewidth]{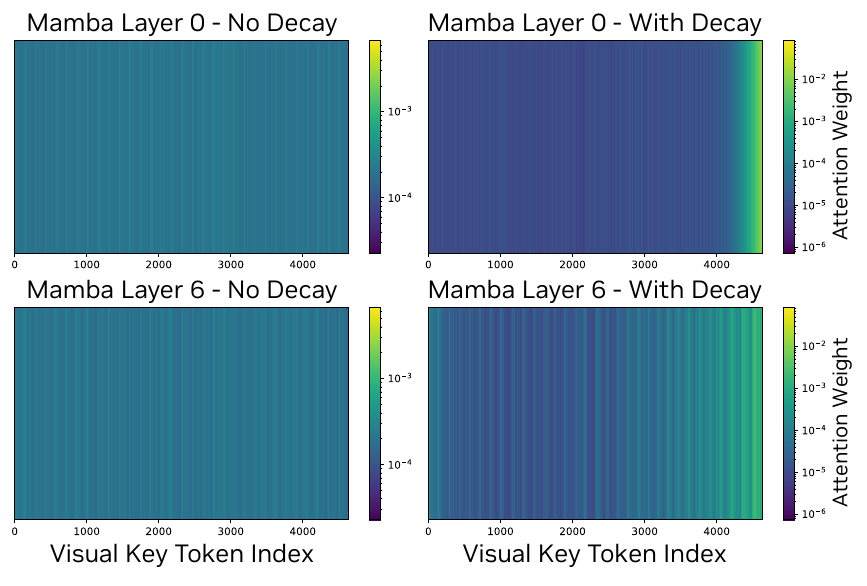}
  \caption{Token Importance Heatmaps W/ and W/O the Decay Term on Two Mamba Layers. Left: using only the attention score $|\bar{\vb}_j^\top \vc_t|$ without decay (\cref{eq:ssm_importance}) produces a more distributed importance pattern across the entire sequence. Right: following Mamba's full implicit attention pattern $|w_{t,j}|$ (\cref{eq:ssm_weight}), these two layers concentrate the importance heavily on the most recent tokens. A logarithm color space is used in this figure.}
  \label{fig:cross_attn_heatmap}
\end{figure}



Based on the mapping established above, we identify $\bar{\vb}_j^\top \vc_t$ (where $\bar{\vb}_j = \Delta_j \vb_j$) as a direct measure of query-visual alignment, analogous to the query-key dot product in attention. In our setting, where the input sequence consists of visual tokens followed by text (query) tokens, $\vc_t$ is computed from the text token at position $t$, while $\bar{\vb}_j$ is computed from the visual token at position $j$ with the gating factor $\Delta_j$ absorbed.

\paragraph{Empirical Inspection of the Decay Effect.}
\cref{fig:cross_attn_heatmap} visualizes token importance computed with and without the decay term across Mamba layers. When following Mamba's full implicit attention pattern (\cref{eq:mamba_implicit_attention}), the cumulative decay $\alpha_{t,j} = \prod_{u=j+1}^{t} \bar{\mA}_u$ causes certain layers' importance scores to concentrate heavily on tokens near the end of the visual sequence. This bias occurs when the diagonal elements of $\bar{\mA}_u$ (lie in $(0,1)$) have relatively small values, causing the weights to decay exponentially with the distance between positions $j$ and $t$. Consequently, early visual tokens receive near-zero importance regardless of their actual content relevance to the query.

In contrast, when we compute importance using only the attention score component $\bar{\vb}_j^\top \vc_t$ without decay, the resulting heatmap shows a more distributed importance pattern across the entire sequence. Therefore, for token selection, we use the importance score:
\begin{equation}
s_i^{(\ell, \text{ssm})} = \frac{1}{MG} \sum_{m,g} \left| \bar{\vb}_i^{(g)\top} \vc_m^{(g)} \right|,
\label{eq:ssm_importance_appendix}
\end{equation}
averaging over $M$ query (text) positions and $G$ groups.

\subsection{Details of the Experimental Setup}
\label{sec:appendix_experimental_setup}
\paragraph{Models.}
We adopt the pretrained SigLIP2 vision encoder~\cite{tschannen2025siglip} with patch size 16 at 384$\times$384 resolution, producing 576 patch tokens. 
An MLP projector then pools every four adjacent patches, yielding 144 visual tokens per frame.
For the LLM stage, we study two architecture families: a dense Transformer and a Mamba--Transformer hybrid. For the Transformer setting, we use the pretrained Qwen3-VL 8B~\cite{bai2025qwen3vltechnicalreport}, built on a 36-layer Qwen3 Transformer with grouped-query attention. 
For the hybrid setting, we use Nemotron-Nano-V2 VL 12B~\cite{nvidia2025nvidianemotronnanov2}, following the Nemotron-H hybrid~\cite{nvidia2025nemotronhfamilyaccurateefficient} design that interleaves attention with Mamba-2 and FFN (MLP) blocks. Specifically, the 12B base architecture consists of 62 layers in total, with 28 Mamba-2, 28 FFN, and 6 self-attention layers (roughly 8\%) evenly dispersed throughout the network. This setup enables us to study token reduction across both attention and Mamba layers and to compare its impact between Transformer and hybrid models. For both architectures, we reuse only the LLM backbones, integrate them with the SigLIP2 vision encoder and the random initialized vision-language projector and apply the identical multi-stage training recipe for both models to obtain the VLM models.

\paragraph{Training.}
Our training consists of four stages: (1) alignment (Stage~0), (2) SFT (Stage~1), (3) long-context finetuning (Stage~2), and (4) long-context finetuning with token-reduction (Stage~3). The maximum number of frames per sample increases from 32 (Stage~1) to 64 (Stage~2) and 128 (Stage~3) to progressively adapt the model to long video inputs; the no-reduction baseline uses 96 frames in the final stage (without reduction), which is the largest that fits in GPU memory.
Following~\cite{jiang2025storm}, Stage~0 uses 95K image--text pairs to align visual tokens to LLM by training only the MLP projector (vision encoder and LLM frozen). Stage~1 performs supervised finetuning (SFT) on a large, diverse mixture of $\sim$12.5M text/image/video examples (up to 32 frames). Stage~2 performs long-context finetuning on the LLaVA-Video dataset~\cite{zhang2024video} (up to 64 frames). Finally, Stage~3 briefly finetunes the model with token reduction on long videos (up to 128 frames) using EAGLE-Video-110K~\cite{chen2025eagle}, a dataset for long-video understanding that includes videos up to 3.5 hours. All stages are trained using 128 NVIDIA A100 80GB GPUs. Alignment, SFT, and long-context fine-tuning take 46 minutes, 11 days, and 25 hours, respectively.

\paragraph{Evaluation Benchmarks.}
We evaluate all token reduction techniques on three long-context video benchmarks: VideoMME~\cite{fu2025video}, LongVideoBench (LongVB)~\cite{wu2024longvideobench}, and LVBench~\cite{wang2025lvbench}. 
VideoMME provides broad video QA; we report VideoMME in the \textit{w/o subtitles} setting to isolate visual-temporal reasoning without relying on transcript cues. LongVB evaluates QA over long clips (up to $\sim$1 hour), while LVBench targets substantially longer videos (30\,min--2\,hrs), further emphasizing long-horizon reasoning. 

\subsection{Reduction Schedule Configurations}
\label{sec:appendix_reduction_schedules}


\begin{table*}[t]
\centering
\small
\renewcommand{\arraystretch}{1.35}

\resizebox{\textwidth}{!}{%
\begin{tabularx}{\textwidth}{l|c|c|c|>{\raggedright\arraybackslash}X}
\hline
\makecell[l]{\textbf{Layer}\\\textbf{Type}} &
\makecell{\textbf{Schedule}} &
\makecell{\textbf{Reduction}\\\textbf{Layers}} &
\makecell{\textbf{Comp.}\\\textbf{(\%)}} &
\makecell[l]{\textbf{Parameters}} \\
\hline

\multicolumn{5}{l}{\textit{Attn Only:} Token reduction applied only at the 6 attention layers at positions 7, 16, 25, 34, 43, 52} \\
\hline
\multirow{4}{*}{\makecell[l]{Attn Only}}
& \multirow{4}{*}{Step Decay}
& 1st Attn & 24.6 & $\alpha_{0{:}6}{=}1.0$; $\alpha_{7{:}61}{=}0.15$ \\
\cline{3-5}
& & \multirow{3}{*}{All Attn}
& 25.2 & $\alpha_{0{:}6}{=}1.0$; $\alpha_{7{:}15}{=}0.25$, $\alpha_{16{:}33}{=}0.20$, $\alpha_{34{:}61}{=}0.10$ \\
\cline{4-5}
& & & 34.7 & $\alpha_{0{:}6}{=}1.0$; $\alpha_{7{:}15}{=}0.50$, $\alpha_{16{:}24}{=}0.40$, $\alpha_{25{:}33}{=}0.30$, $\alpha_{34{:}42}{=}0.20$, $\alpha_{43{:}61}{=}0.10$ \\
\cline{4-5}
& & & 50.1 & $\alpha_{0{:}6}{=}1.0$; $\alpha_{7{:}15}{=}0.65$, $\alpha_{16{:}24}{=}0.60$, $\alpha_{25{:}33}{=}0.50$, $\alpha_{34{:}42}{=}0.40$, $\alpha_{43{:}51}{=}0.30$, $\alpha_{52{:}61}{=}0.20$ \\
\hline

\multicolumn{5}{l}{\textit{Mamba Only:} Token reduction only applied once at one early Mamba layer} \\
\hline
\multirow{2}{*}{\makecell[l]{Mamba Only}}
& \multirow{2}{*}{Step Decay}
& 1st Mamba & 25.0 & $\alpha_{0{:}61}{=}0.25$ \\
\cline{3-5}
& & 2nd Mamba & 25.5 & $\alpha_0{=}1.0$; $\alpha_{1{:}61}{=}0.23$ \\
\hline

\multicolumn{5}{l}{\textit{Mamba+Attn (Step Decay):} Interleaving reduction with both attention layers and Mamba layers} \\
\hline
\multirow{2}{*}{\makecell[l]{Mamba+Attn}}
& \multirow{2}{*}{Step Decay}
& All Attn+1M & 25.4 & $\alpha_0{=}1.0$; $\alpha_{1{:}33}$: $0.32{\to}0.15$ (6 attn + 1 mamba between each attn pair) \\
\cline{3-5}
& & All Attn+2M & 25.4 & $\alpha_0{=}1.0$; $\alpha_{1{:}33}$: $0.32{\to}0.15$ (6 attn + 2 mamba between each attn pair) \\
\hline

\multicolumn{5}{p{\dimexpr\textwidth-2\tabcolsep\relax}}{%
\textit{Mamba+Attn (Sigmoid):} All-layer reduction with sigmoid schedule
$\alpha_\ell = \alpha_{\text{end}} + (\alpha_{\text{start}} - \alpha_{\text{end}}) \cdot \sigma(k(x_0 - \ell/L))$.} \\
\hline
\multirow{3}{*}{\makecell[l]{Mamba+Attn}}
& \multirow{3}{*}{Sigmoid}
& \multirow{3}{*}{All}
& 25.1 & $k{=}20$, $x_0{=}0.11$; $\alpha_{\text{start}}{=}1.0$, $\alpha_{\text{end}}{=}0.125$ \\
\cline{4-5}
& & & 35.0 & $k{=}20$, $x_0{=}0.24$; $\alpha_{\text{start}}{=}1.0$, $\alpha_{\text{end}}{=}0.125$ \\
\cline{4-5}
& & & 50.2 & $k{=}20$, $x_0{=}0.41$; $\alpha_{\text{start}}{=}1.0$, $\alpha_{\text{end}}{=}0.125$ \\
\hline

\end{tabularx}
}

\caption{Complete reduction schedule configurations for hybrid models (Nemotron-Nano-V2 VL 12B). Following the official model configuration~\cite{nvidia2025nvidianemotronnanov2}, we include layer indices for MLP layers when explaining the reduction parameters. \textbf{Layer Type} indicates the layer types performing token reduction: ``Attn Only'' applies reduction only at the 6 attention layers; ``Mamba Only'' includes reduction starting from a single Mamba layer; ``Mamba+Attn'' applies reduction at a subset or all 62 layers (excluding MLP layers). \textbf{Reduction Layers} specifies the layer subset: ``1st Attn'' reduces tokens starting from the first attention layer; ``All attn'' applies different rates at each of the 6 attention layers; ``1st Mamba'' and ``2nd Mamba'' for single Mamba layer configurations; ``All Attn+1M'' and ``All Attn+2M'' for 6 attention layers plus 1 or 2 Mamba reduction layers between each attention pair; ``All'' for all layers with progressive reduction. Layer indices are 0-indexed. All configurations enforce a minimum of 144 tokens to prevent reduction on single image data.}
\label{tab:reduction_schedule_details}
\end{table*}

\cref{tab:reduction_schedule_details} details the reduction schedule configurations used in our experiments on the Nemotron-Nano-V2 VL 12B hybrid model~\citep{nvidia2025nvidianemotronnanov2}. The model consists of 62 layers total (30 Mamba + 30 MLP + 6 attention), where attention layers appear at positions 7, 16, 25, 34, 43, and 52 (0-indexed). In the table, $\alpha_\ell$ denotes the token retention ratio at layer $\ell$, i.e., the fraction of tokens kept after reduction (e.g., $\alpha{=}0.25$ means retaining 25\% of tokens). We employ two schedule types:

\paragraph{Sigmoid Schedule.} The retention rate $\alpha_\ell$ at layer $\ell$ follows a sigmoid function:
\begin{equation}
\alpha_\ell = \alpha_{\text{end}} + (\alpha_{\text{start}} - \alpha_{\text{end}}) \cdot \sigma\bigl(k(x_0 - \ell/L)\bigr),
\end{equation}
where $\sigma(\cdot)$ is the sigmoid function, $L$ is the total number of layers, $k$ controls the steepness of the transition, and $x_0$ determines the midpoint point (as a fraction of total layers). A smaller $x_0$ delays reduction to later layers, implementing the low-to-high pattern that preserves more tokens early when importance scores are less reliable. Larger $k$ produces sharper transitions between high and low retention regions. In our experiments, we fix the steepness $k=20$ and the start value $\alpha_{\text{start}}{=}1.0$ (no reduction initially) and the end value $\alpha_{\text{end}}{=}0.125$ (retain 12.5\% at final layers), varying $x_0 \in \{0.11, 0.24, 0.41\}$ to achieve compression rates of approximately 25\%, 35\%, and 50\% respectively. The sigmoid schedule provides smooth, progressive transitions of the reduction rate across all layers.

\paragraph{Step Decay Schedule.} The retention rate $\alpha_\ell$ is specified explicitly for groups of layers, allowing fine-grained control over reduction patterns. For attention-only reduction, we define rates for the 6 attention positions. For example, $\alpha_{0:6}{=}1.0$; $\alpha_{7:61}{=}0.15$ means layers 0--6 retain all tokens while layers 7--61 (i.e., after the first attention layer) retain 15\%. For hybrid reduction spanning multiple layer types, we typically set $\alpha_0{=}1.0$ (no reduction at the first layer) and gradually decay across subsequent reduction layers. We hypothesize that this low-to-high sparsity levels allow the early Mamba layer to compress as much information as possible in the hidden state before the later layers starts to prune tokens. This pattern empirically yields better performance than uniform reduction variants, e.g., 1st Mamba or 2nd Mamba variants.

\paragraph{Layer Types.} We evaluate three strategies for where to apply reduction: (1) \emph{Attention-only}, reducing tokens only at the 6 attention layers; (2) \emph{Mamba-only}, reducing at early Mamba layers but maintain a relatively low sparsity on deeper layers; and (3) \emph{Hybrid (Mamba+Attn)}, applying reduction at both attention and selected Mamba layers. For hybrid schedules, the notation ``All Attn+1M'' indicates 1 Mamba reduction between each pair of attentions, so there are 6 attention layers plus 7 Mamba layers. Similarly, ``All Attn+2M'' indicates 2 Mamba layers between each pair of attentions. All configurations enforce a minimum of 144 tokens to preserve full tokens for image data in training.


\begin{table*}[htbp]
\centering
\scriptsize
\setlength{\tabcolsep}{4pt}
\renewcommand{\arraystretch}{1.45}

\definecolor{hlrow}{RGB}{248,216,216} 
\definecolor{posblue}{RGB}{0,60,200}
\definecolor{negred}{RGB}{200,0,0}

\newcommand{\trtpos}[1]{\textcolor{posblue}{#1}}
\newcommand{\trtneg}[1]{\textcolor{negred}{#1}}
\newcommand{\trtvaldelta}[2]{%
  \begingroup
  \edef\trtdelta{\fpeval{#1-(#2)}}%
  \mbox{\num{#1}\,%
    \ifdim\trtdelta pt=0pt
      \trtpos{(+\num{0})}%
    \else
      \ifdim\trtdelta pt>0pt
        \trtpos{(+\num{\trtdelta})}%
      \else
        \trtneg{(-\num{\fpeval{-(\trtdelta)}})}%
      \fi
    \fi}%
  \endgroup
}
\newcommand{\trtttftfactor}[2]{%
  \begingroup
  \edef\trtvalue{\fpeval{round(#1,2)}}%
  \edef\trtfactor{\fpeval{round((#2)/(#1),1)}}%
  \mbox{\trtvalue\,%
    \ifdim\trtfactor pt>1pt
      \trtpos{(\texttimes\trtfactor)}%
    \else
      \ifnum\pdfstrcmp{\trtfactor}{1}=0
        \trtpos{(\texttimes 1.00)}%
      \else
        \trtneg{(\texttimes\trtfactor)}%
      \fi
    \fi}%
  \endgroup
}

\def\trtBaseVMME{69.89}
\def\trtBaseLVB{66.64}
\def\trtBaseLV{50.94}
\def\trtBaseAVG{62.49}
\def\trtBaseTTFT{4.78}

\newcolumntype{Y}{>{\centering\arraybackslash}X}
\begin{tabularx}{\textwidth}{l|Y|Y|Y|Y|Y|Y}
\hline
\makecell[l]{\textbf{Token}\\\textbf{Reduction}} &
\makecell{\textbf{Comp. Rate}\\\textbf{(\%)}} &
\makecell{\textbf{VideoMME}\\\textbf{(1$\sim$60m)}} &
\makecell{\textbf{LongVB}\\\textbf{(8s$\sim$60m)}} &
\makecell{\textbf{LVBench}\\\textbf{(30m$\sim$2h)}} &
\textbf{Avg} &
\textbf{TTFT (s)}\\
\hline

\hline
\rowcolor{black!6}
Baseline & 100 & {69.89} & {66.64} & {50.94} & {62.49} & 4.78 \\
\hline

\multicolumn{7}{l}{\textbf{Train-Time Reduction}} \\
\hline

1st layer & 25.0 &
\trtvaldelta{59.52}{\trtBaseVMME} &
\trtvaldelta{55.95}{\trtBaseLVB} &
\trtvaldelta{38.28}{\trtBaseLV} &
\trtvaldelta{51.25}{\trtBaseAVG} &
\trtttftfactor{1.07}{\trtBaseTTFT}\\


\rowcolor{hlrow}
6 layers & 25.1 &
\trtvaldelta{68.41}{\trtBaseVMME} &
\textbf{\trtvaldelta{64.62}{\trtBaseLVB}} &
\trtvaldelta{43.19}{\trtBaseLV} &
\trtvaldelta{58.74}{\trtBaseAVG} &
\trtttftfactor{1.18}{\trtBaseTTFT}\\

13 layers & 25.1 &
\trtvaldelta{67.93}{\trtBaseVMME} &
\trtvaldelta{63.65 }{\trtBaseLVB} &
\trtvaldelta{44.29}{\trtBaseLV} &
\trtvaldelta{58.62}{\trtBaseAVG} &
\trtttftfactor{1.09}{\trtBaseTTFT}\\

20 layers & 25.0 &
\trtvaldelta{68.00}{\trtBaseVMME} &
\trtvaldelta{63.87 }{\trtBaseLVB} &
\trtvaldelta{44.16}{\trtBaseLV} &
\trtvaldelta{58.68}{\trtBaseAVG} &
\trtttftfactor{1.09}{\trtBaseTTFT}\\

\rowcolor{hlrow}
All & 25.5 &
\textbf{\trtvaldelta{68.52}{\trtBaseVMME}} &
\trtvaldelta{64.32}{\trtBaseLVB} &
\textbf{\trtvaldelta{45.45}{\trtBaseLV}} &
\textbf{\trtvaldelta{59.43}{\trtBaseAVG}} &
\trtttftfactor{1.17}{\trtBaseTTFT}\\
\hline
\end{tabularx}
\caption{\textbf{Token Reduction with Different Reduction Patterns for Qwen3-VL 8B.} Train-time reduction is applied at selected LLM layers (1st, 6/13/20, or all), using the same relative layer positions as the hybrid setup. 
``Comp. Rate" is remaining tokens; 
($\cdot$) shows change from baseline. 
TTFT is LLM-stage latency on a single A100 with a 256-frame input. 
}
\label{tab:token_reduction_transformer}
\end{table*}

\begin{figure*}[t]
  \begin{subfigure}[t]{0.49\textwidth}
  \centering
  \includegraphics[width=1.0\linewidth]{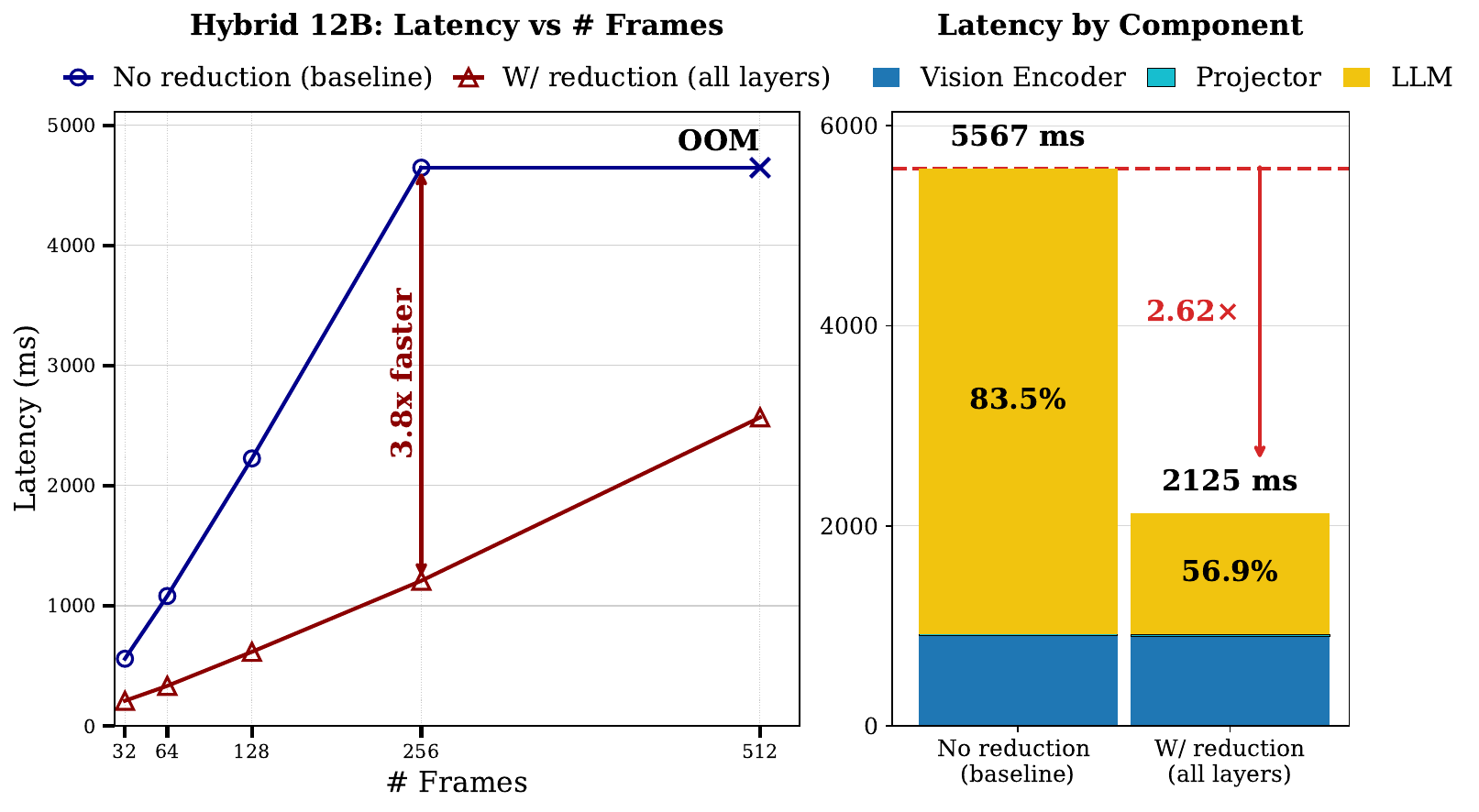}
  \caption{\textbf{Nemotron-Nano-V2 VL 12B (hybrid).}}
  \label{fig:latency_hybrid}
  \end{subfigure}
  \hfill
  \begin{subfigure}[t]{0.49\textwidth}
  \centering
  \includegraphics[width=1.0\linewidth]{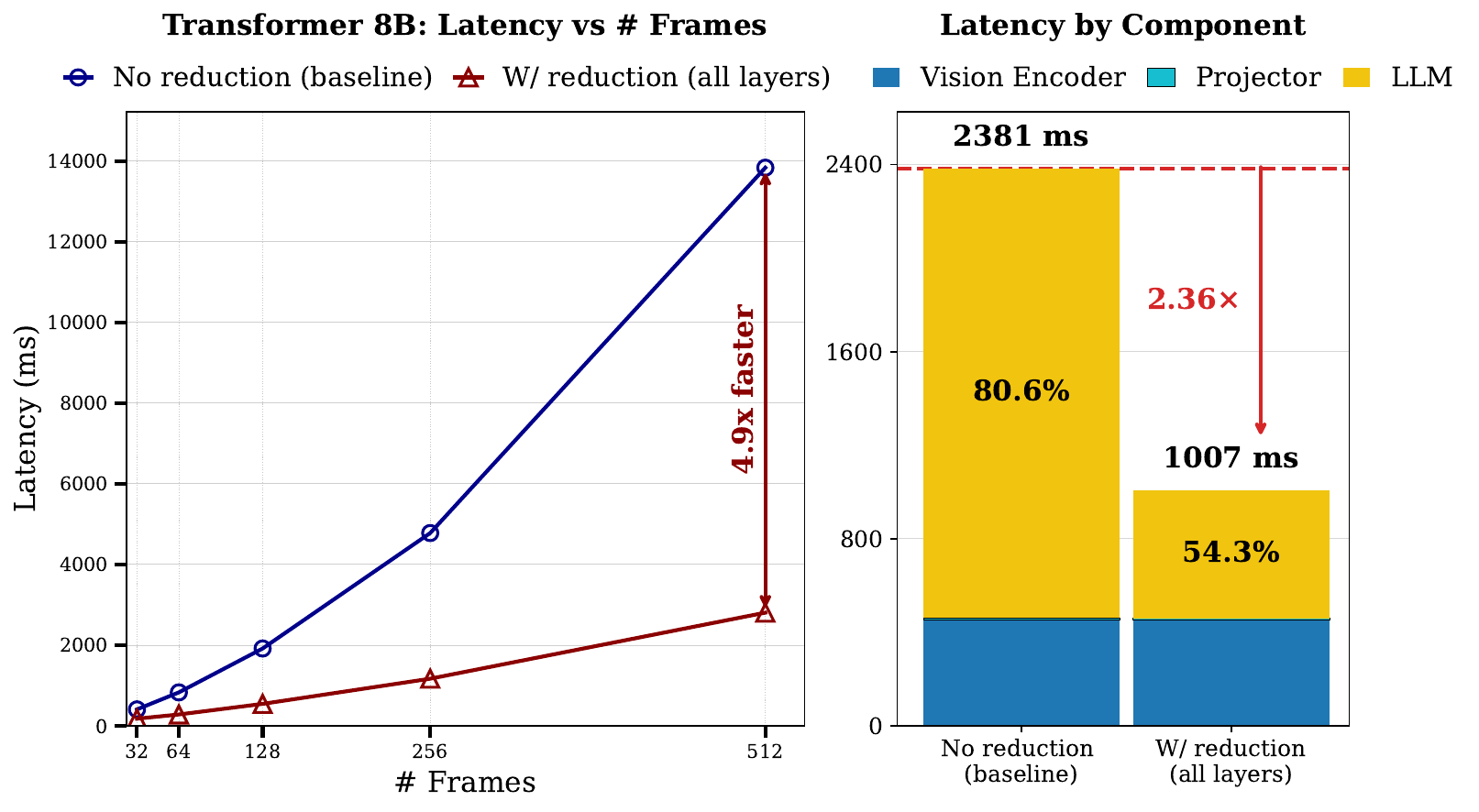}
  \caption{\textbf{Qwen3-VL 8B (Transformer).}}
  \label{fig:latency_transformer}
  \end{subfigure}
  \caption{\textbf{Latency Analysis.} \textbf{(left)} LLM-stage latency vs.\ frames on a single A100, with/without token reduction. \textbf{(right)} Component-wise latency (vision encoder / projector / LLM) on a 128-frame VideoMME input.}
\end{figure*}


\begin{table*}[ht]
\centering
\scriptsize
\setlength{\tabcolsep}{5pt}
\renewcommand{\arraystretch}{1.3}
\begin{tabular}{l|l|c|c|c|c}
\hline
\multirow{2}{*}{\textbf{Model Type}} & \multirow{2}{*}{\textbf{Model / Method}} & \multirow{2}{*}{\textbf{Size}} &
\multicolumn{1}{c|}{\textbf{VideoMME}} & \multicolumn{1}{c|}{\textbf{LongVB}} & \multicolumn{1}{c}{\textbf{LVBench}} \\
  &  &  & (1$\sim$60m) & (8s$\sim$60m) & (30m$\sim$2h) \\
\hline

\multicolumn{5}{l}{\textbf{Proprietary Models}} \\
\hline
 \multirow{2}{*}{Transformer} & GPT-4o~\cite{hurst2024gpt} & -- & 71.9 & 66.7 & 34.7 \\
  & Gemini-1.5-Pro~\cite{team2023gemini} & -- & 75.0 & 64.0 & 33.1 \\
\hline


\multicolumn{6}{l}{\textbf{Uniform Token Selection}} \\
\hline
 \multirow{6}{*}{Transformer}  & LLaVA-Onevision~\cite{li2024llava} & 7B & 58.2 & 56.4 & -- \\
 & Qwen2-VL~\cite{wang2024qwen2} & 7B & 63.3 & 55.6 & 42.4 \\
 & NVILA~\cite{liu2025nvila} & 8B & 64.2 & 57.7 & -- \\
 & Apollo~\cite{zohar2025apollo} & 7B & 61.3 & 58.5 & -- \\
 & VideoLLaMA3~\cite{zhang2025videollama} & 7B & 66.2 & 59.8 & 45.3 \\
 & Oryx-1.5~\cite{liu2024oryx} & 34B & 67.3 & 62.0 & 30.8 \\
\hline

\multicolumn{6}{l}{\textbf{Adaptive Token Selection/Reduction}} \\
\hline
 \multirow{8}{*}{Transformer}  & MLLM-FS~\cite{hu2025m} & 7B & 58.7 & 57.0 & -- \\
 & LongVU~\cite{shen2024longvu} & 7B & 60.6 & -- & -- \\
 & SF-LLaVA-1.5~\cite{xu2025slowfast} & 7B & 63.9 & 62.5 & 45.3 \\
 & ViLAMP~\cite{cheng2025scaling} & 7B & 67.5 & 61.2 & 45.2 \\
 & {Qwen2.5 VL+SparseVILA}~\cite{khaki2025sparsevila} & 7B & 66.3 & 60.1 & -- \\
 & Qwen2.5 VL+FlexSelect~\cite{zhang2025flexselect} & 7B & 68.2 & 62.4 & 51.2 \\
 & LLaVA-Video+FlexSelect~\cite{zhang2025flexselect} & 7B & 68.9 & 61.9 & 52.9 \\
\cline{2-6}


 & Qwen2.5+FlexSelect~\cite{zhang2025flexselect} & 72B & \textbf{74.4} & \textbf{66.4} & \textbf{56.6} \\
\hline
\multicolumn{6}{l}{\textbf{Our Token Reduction}} \\
\hline
\rowcolor{black!6}
\rowcolor{hlrow}
Transformer & Qwen3 VL + All layer reduction & 8B & 68.52 & 64.32 & 45.45 \\
\rowcolor{black!6}
\rowcolor{hlrow}
Hybrid & Nemotron-Nano-V2 VL + All layer reduction & 12B & \textbf{69.70} & \textbf{66.04} & \textbf{54.29} \\
\hline

\end{tabular}
\caption{\textbf{Comparison on Long-Video Benchmarks.} We report scores on VideoMME (w/o subtitles), LongVB (LongVideoBench; 8s$\sim$60m), and LVBench (30m$\sim$2h) for proprietary, open-source LLMs with uniform/adaptive token selection, and our method. Ours use all-layer reduction for Qwen3-VL (Transformer) and Nemotron-Nano-V2 VL (Hybrid).
}
\label{tab:comparison}
\end{table*}

\subsection{End-to-End Latency Across Frame Counts}
\label{sec:appendix_e2e_latency}

\cref{tab:e2e_latency} reports end-to-end inference latency (vision encoder + projector + LLM, in seconds) for Nemotron-Nano-V2 VL 12B on a single NVIDIA A100, sweeping the number of input frames from $16$ to $256$. We compare three configurations: no reduction, attention-only token reduction at $\sim$25\% token budget, and the full attention+Mamba (\texttt{All}-layer) reduction at $\sim$25\%. The all-layer variant achieves a $1.6\times$ speedup at $16$ frames, growing to $2.4$--$2.6\times$ for $64$--$256$ frames. The relative gain widens with input length because the LLM prefilling cost grows superlinearly while the vision encoder cost scales linearly with the number of frames; for long videos the LLM stage dominates total runtime, so its acceleration translates almost directly into the end-to-end metric. This complements \cref{fig:end_to_end_latency_hybrid} (LLM-stage TTFT only) by accounting for the full prefill pipeline.

\begin{table*}[t]
\centering
\footnotesize
\setlength{\tabcolsep}{4pt}
\renewcommand{\arraystretch}{1.25}
\begin{tabular}{l|c|c|c|c|c}
\hline
\textbf{Frames} & \textbf{16} & \textbf{32} & \textbf{64} & \textbf{128} & \textbf{256}\\
\hline
No Reduction (100\%)        & 0.349 & 0.677 & 1.313 & 2.714 & 5.567 \\
Attention-only (25\%)       & 0.179 & 0.287 & 0.520 & 1.033 & 2.052 \\
Attn+Mamba \texttt{All} (25\%) & 0.214 & 0.327 & 0.564 & 1.080 & 2.125 \\
\hline
Speedup (Attn+Mamba vs.\ baseline) & $1.6\times$ & $2.1\times$ & $2.3\times$ & $2.5\times$ & $2.6\times$ \\
\hline
\end{tabular}
\caption{\textbf{End-to-End Inference Latency} for Nemotron-Nano-V2 VL 12B on a single NVIDIA A100, in seconds (vision encoder $+$ projector $+$ LLM). Token reduction at a $\sim$25\% budget yields a $1.6\times$ speedup at $16$ frames and $2.4$--$2.6\times$ speedup at $64$--$256$ frames; the gap widens with frame count because LLM prefilling cost grows super-linearly while vision encoder cost is linear.}
\label{tab:e2e_latency}
\end{table*}

\subsection{Training-Time Efficiency}
\label{sec:appendix_training_efficiency}

Token reduction is not only an inference-time optimization. We also benchmark training throughput and memory under \texttt{All}-layer reduction at $\sim$25\% budget against no-reduction training, with $64$-frame video samples and otherwise identical training resources. The hybrid model's training run time drops from $1251.7$\,s to $862.3$\,s ($1.45\times$ speedup), wall-clock time drops from $1390$\,s to $942$\,s ($1.48\times$), and throughput increases from $0.818$ to $1.187$ samples/sec. Peak GPU memory drops from $77{,}436$\,MiB to $70{,}664$\,MiB, a saving of $\sim$$6.6$\,GB ($8.7\%$). On A100~80\,GB, the no-reduction baseline supports at most $96$ frames per sample, while the reduced variant supports $128$ frames in the same memory budget, enabling $33\%$ more visual context per sample. Token reduction is therefore an essential lever for scaling long-video training, not merely inference.

\subsection{Impact Statements}
\label{sec:impact}
This paper presents work whose goal is to advance the field of machine learning by improving the efficiency of long-video vision--language models. The primary intended impact is reduced inference latency and compute/energy cost for long-context video understanding. Beyond this, we do not anticipate additional societal consequences that require specific discussion.

\subsection{Related Works}
\label{sec:related_appendix}

Token reduction is widely used in multimodal LLMs to reduce redundant computation, especially for long video inputs. Token pruning methods are often grouped into query-agnostic approaches (based on visual similarity) and query-aware approaches (conditioned on language). Query-aware methods typically score tokens using text-to-vision attention or cross-modal similarity. FastV~\cite{chen2024imageworth12tokens} prunes tokens using early attention signals, and SparseVLM~\cite{zhang2025sparsevlmvisualtokensparsification} prunes visually irrelevant tokens using cross-attention scores. Instead of pruning, token merging aggregates redundant tokens, such as multi-granular spatio-temporal token merging~\cite{hyun2025multi}. Most of these methods are designed around attention layers and do not explicitly model token importance inside non-attention blocks, which is important for hybrid architectures.

Many works focus on long-video understanding by reducing temporal redundancy. FlexSelect~\cite{zhang2025flexselect}, DynTok~\cite{zhang2025dyntok}, and DyCoke~\cite{tao2025dycoke} select or compress tokens for long videos, and METok~\cite{wang2025metok} proposes multi-stage event-based compression. Our work also targets long-context video, but focuses on hybrid models and uses depth-dependent token-importance behavior to design reduction schedules.

Another distinction is where reduction is applied. Some methods prune once early (before the LLM or at early layers) and keep the reduced set fixed. For example, FastV~\cite{chen2024imageworth12tokens} prunes after early layers, and VisionSelector~\cite{zhu2025visionselector} selects tokens before the LLM. FlexSelect~\cite{zhang2025flexselect} and VISA~\cite{jiang2025visa} also produce a reduced token set for inference. Other methods prune progressively across layers to reduce the sensitivity of early pruning. PyramidDrop~\cite{xing2024pyramiddrop} drops tokens across depth, and LaCo~\cite{liu2025laco}, VisionZip~\cite{yang2025visionzip}, DynTok~\cite{zhang2025dyntok}, and METok~\cite{wang2025metok} apply multi-layer or multi-stage reduction. Our method follows this direction and uses a progressive schedule that keeps more tokens early and prunes more later.

\end{document}